\documentclass[lettersize,journal]{IEEEtran}
\usepackage{amsmath,amsfonts, mathtools, amssymb, multirow}
\usepackage{algorithm, multirow, algpseudocode}
\usepackage{array}
\usepackage{hyperref}
\usepackage{textcomp}
\usepackage{stfloats}
\usepackage{url}
\usepackage{verbatim}
\usepackage{graphicx}
\usepackage{cite}
\usepackage{subcaption, multirow, hyperref}

\hypersetup{
	colorlinks=true,
	linkcolor=blue,
	citecolor=blue,
	urlcolor=blue,
}

\begin{document}

\title{Physics-Informed Reinforcement Learning of Spatial Density Velocity Potentials for Map-Free Racing}

\author{Shathushan Sivashangaran, Apoorva Khairnar, Sepideh Gohari, Vihaan Dutta and Azim Eskandarian
\thanks{\textit{Corresponding author: Shathushan Sivashangaran}}
\thanks{Shathushan Sivashangaran, Sepideh Gohari and Azim Eskandarian are with the Autonomous Robots and Vehicles Laboratory, College of Engineering, Virginia Commonwealth University, Richmond, VA 23284, USA. (email: \href{mailto:sivashangars@vcu.edu}{sivashangars@vcu.edu}; \href{mailto:goharis@vcu.edu}{goharis@vcu.edu}; \href{mailto:eskandariana@vcu.edu}{eskandariana@vcu.edu}).}%
\thanks{Apoorva Khairnar is with the Department of Mechanical Engineering, Virginia Tech, Blacksburg, VA 24061, USA. (email: \href{mailto:apoorvak@vt.edu}{apoorvak@vt.edu}).}
\thanks{Vihaan Dutta is with the Robotics Department, University of Michigan, Ann Arbor, MI 48109, USA. (email: \href{mailto:vihdutta@umich.edu}{vihdutta@umich.edu}).}}%

\markboth{}%
{Sivashangaran \MakeLowercase{\textit{et al.}}: Physics-Informed Reinforcement Learning of Spatial Density Velocity Potentials for Map-Free Racing}

\maketitle

\begin{abstract}

Autonomous racing without prebuilt maps is a grand challenge for embedded robotics that requires kinodynamic planning from instantaneous sensor data at the acceleration and tire friction limits. Out-Of-Distribution (OOD) generalization to various racetrack configurations utilizes Machine Learning (ML) to encode the mathematical relation between sensor data and vehicle actuation for end-to-end control, with implicit localization. These comprise Behavioral Cloning (BC) that is capped to human reaction times and Deep Reinforcement Learning (DRL) which requires large-scale collisions for comprehensive training that can be infeasible without simulation but is arduous to transfer to reality, thus exhibiting greater performance than BC in simulation, but actuation instability on hardware. This paper presents a DRL method that parameterizes nonlinear vehicle dynamics from the spectral distribution of depth measurements with a non-geometric, physics-informed reward, to infer vehicle time-optimal and overtaking racing controls with an Artificial Neural Network (ANN) that utilizes less than 1\% of the computation of BC and model-based DRL. Slaloming from simulation to reality transfer and variance-induced conservatism are eliminated with the combination of a physics engine exploit-aware reward and the replacement of an explicit collision penalty with an implicit truncation of the value horizon. The policy outperforms human demonstrations by 12\% in OOD tracks on proportionally scaled hardware, by maximizing the friction circle with tire dynamics that resemble an empirical Pacejka tire model. System identification illuminates a functional bifurcation where the first layer compresses spatial observations to extract digitized track features with higher resolution in corner apexes, and the second encodes nonlinear dynamics. 

\end{abstract}

\begin{IEEEkeywords}
Autonomous Racing, Kinodynamic Planning, Reinforcement Learning, Simulation Training, System Identification
\end{IEEEkeywords}

\section{Introduction} \label{se:introduction}

\IEEEPARstart{A}utonomous racing entails perception, planning, and control systems that operate at the boundaries of vehicle handling at high speeds. Conventional approaches rely predominantly on detailed prebuilt maps, global reference trajectories, and computationally expensive optimal solvers \cite{betz2022autonomous}. While effective with exact states, these map dependent architectures are fundamentally brittle in the real world, where localization errors cascade into the state vector, inhibiting performance. Moreover, this intrinsic dependence on prior map and trajectory knowledge inherently prevents effective Out-Of-Distribution (OOD) generalization, as the control policies are inextricably tied to the specific mathematically optimized geometries.

Machine Learning (ML) circumvents this by computing the mathematical relation between sensor observations and nonlinear vehicle actuation, facilitating end-to-end control with implicit localization to minimize error propagation, conditioned over a large number of training data samples. End-to-end architectures utilizing Behavioral Cloning (BC) successfully deploy on hardware and generalize to different track layouts by training directly on human demonstrations \cite{zarrar2024tinylidarnet}. Although these BC models demonstrate high sample efficiency, training on around ten thousand real world demonstration samples, compared to the millions of simulation steps required by recent Deep Reinforcement Learning (DRL) works \cite{ghignone2025rlpp, evans2023comparing, brunnbauer2022latent}, the reliance on human datasets enforces a strict performance ceiling. DRL models, unbounded by human reaction times, are capable of lapping more than twice as fast as BC in simulation \cite{bosello2022train, evans2023comparing}, and achieve superhuman performance in driving video games \cite{fuchs2021super}.

The scale of DRL training data, and the trial-and-error experience rollout that necessitates collisions, without a safety supervisor, to parameterize time-optimal kinodynamic planning \cite{evans2022accelerating}, render real world training arduous. While these policies demonstrate success in simulation, the methods collapse upon physical deployment, resulting in actuation instability and rapid slaloming \cite{evans2023comparing, brunnbauer2022latent}. To overcome the simulation to reality (sim-to-real) gap, contemporary DRL techniques utilize trajectory tracking hybrid residual architectures \cite{ghignone2025rlpp} that correct geometric tracking errors of a baseline controller such as Pure Pursuit (PP), requiring prebuilt maps and a precomputed optimized racing line.

Precomputed mathematically optimal reference trajectories \cite{heilmeier2020minimum, sivashangaran2022nonlinear}, such as those generated with classical minimum curvature solvers, are commonly used in the DRL reward formulation. While these serve as rigorous benchmarks for cross comparison, utilizing these geometries as explicit training targets inherently restricts the learning process, as the agent is compelled into a regime of pure geometric path following that prevents the encoding of fundamental relations between spatial observations and nonlinear vehicle dynamics. Evaluations of DRL methods that comprise full planning, trajectory tracking, and end-to-end techniques corroborate this limitation \cite{evans2023comparing}. The latter, unlike the others, does not require track information that necessitates a prebuilt map, thus, while it does not achieve the fastest lap time in the training track, generalizes to new tracks with the most favorable sim-to-real transfer. Among the methods that utilize track information, full planning uses the centerline points whereas trajectory tracking minimizes deviation from an optimal reference trajectory, hence the latter is fastest in the training track but generalizes the worst. 

The catalyst for the reality gap is the reliance on low-fidelity, open-source simulation tools. Development of DRL racing architectures predominantly utilize the F1TENTH simulator \cite{o2020f1tenth, charles2025advancing}, which expedited racing ML research. This comprises a 2D kinematics and dynamics solver that does not simulate 3D rigid body interactions, thus policies overfit to a simplified mathematical ideal. Furthermore, utilizing 3D physics engines does not guarantee sim-to-real performance \cite{brunnbauer2022latent}. This dichotomy illustrates the severity of the sim-to-real gap. While autonomous DRL agents \cite{evans2023comparing, bosello2022train, brunnbauer2022latent} achieve lap times significantly faster than State-Of-The-Art (SOTA) end-to-end BC models \cite{zarrar2024tinylidarnet}, the policies lose performance when transferred to reality. 

This paper resolves these bottlenecks by formulating a physics-informed reward within an analytically accurate physics engine to parameterize nonlinear dynamics from spectral spatial density velocity potentials, rather than geometric occupancy, interpreted from an instantaneous array of depth measurements, in an efficient Artificial Neural Network (ANN) with a map-and-model-free DRL policy, that requires less than 1\% of the compute of BC and model-based DRL. By explicitly encoding nonlinear dynamics with a low-speed training environment, our approach achieves the stabilizing benefits of curriculum training, without the requirement for intermediate speed scheduling, enabling zero-shot transfer of encoded physics to high-speed domains. To bridge the sim-to-real gap without inducing actuation instability which is the primary performance inhibitor in sim-to-real works, we introduce an auxiliary physics engine exploit-aware reward that eliminates slaloming by addressing a root instability of discrete-time physics engines, which may mathematically permit instantaneous transitions between extreme actuator limits \cite{brunnbauer2022latent}. Furthermore, we treat collisions as an implicit value truncation without an explicit penalty. This subtle shift removes the mathematical prioritization of crash minimization, encouraging thorough exploration of the friction circle and improved generalization to new track layouts, assessed via reward ablations.

The success of this framework is predicated on an expansion of environmental interaction scale. By extending training to 20,000,000 steps, the agent experiences a manifold of 15,747 edge case collisions. Notably, despite not utilizing an optimal reference trajectory to guide training, our Proximal Policy Optimization (PPO) \cite{schulman2017proximal} agent achieves a mean lateral deviation of $0.08\%$ from the optimal minimum curvature path \cite{heilmeier2020minimum} in the training environment. Furthermore, this geometry independent encoding transfers zero-shot to OOD tracks, maintaining collision-free laps and exhibiting sophisticated racing line anticipation, with an average deviation of $14.60\%$ from the mathematical optimum. Moreover, the policy transfers zero-shot to proportionally scaled hardware \cite{sivashangaran2023xtenth} to lap 26\% faster than a Nonlinear Predictive Geometric Proportional-Integral-Derivative (PID) controller that was the fastest method to qualify with 10 collision-free laps at the 2023 IEEE Intelligent Vehicles Symposium, and 12\% faster than a human demonstration, that represents a best case BC performance, in OOD tracks.

To further demonstrate the efficacy of the spectral velocity
potential encoding for dynamics-optimized kinodynamic plan-
ning, we apply the same formula to parameterize overtake
maneuvers. Without altering the core Reinforcement Learning (RL) formulation, the policy infers optimized high-momentum overtake trajectories around dynamic obstacle vehicles, with post-training in a multi-agent environment, without specialized multi-agent formulations and reference trajectories \cite{trumpp2024racemop}. Moreover, we conduct system identification on the spatial-kinodynamic inference by dissecting the ANN's internal activations to elucidate the internal mechanisms driving this end-to-end framework. By systematically mapping inter-layer correlations, a distinct functional bifurcation is identified, where the initial layer operates as a feature extractor that compresses observations to digitized track features with higher resolution in paramount corner apexes, while the subsequent layer encodes nonlinear dynamics to handle the vehicle at the boundary of the tire friction circle, that better fits a nonlinear empirical Pacejka tire model \cite{bakker1987tyre} than a linear kinematic model \cite{polack2017kinematic}.

In summary, the contributions of this paper are as follows.

\begin{itemize}

\item Spectral Kinodynamic Encoding of Spatial Density Velocity Potentials with a Physics-Informed Reward: The interpretation of instantaneous depth measurements as spectral signals, mapping spatial variance to velocity potentials and dynamic limits via a self-supervised physics-informed throttle-maximization objective, with the stabilizing benefits of a curriculum without intermediate scheduling, to organically teach the agent momentum-conserving, physically viable controls without prescriptive geometric mimicry, sequential temporal data or explicit velocity inputs. The independence of reference trajectories in the reward enables encoding dynamics-optimized overtaking with the same RL formulation.

\item Auxiliary Reward for Sim-to-Real Transfer and Collision Value Truncation: The combination of an auxiliary oscillation reward that penalizes exploitation of discrete-time simulator permitted steering reversals, with an implicit value truncation for collisions that zeroes future expected returns instead of applying a traditional collision termination penalty, to eliminate both sim-to-real slaloming and variance-induced conservatism, ensuring OOD generalization and smooth controls on hardware. 

\item Large-Scale Environmental Interaction: A training paradigm reflecting ANN scaling laws, expanding the environmental interaction to 20,000,000 simulation steps. This order-of-magnitude increase in scale, compared to geometry guided DRL architectures that comprise residual techniques trained in 2,000,000 training steps \cite{ghignone2025rlpp}, and model-free that is inadequate with 8,000,000 necessitating model-based trained in 2,000,000 \cite{brunnbauer2022latent}, allows the agent to experience over 15,747 edge-case collisions, successfully surmounting local optima, converging to the global optimum, and condensing the relations between depth observations and nonlinear rigid-body dynamics into an efficient 2-layer Multilayer Perceptron (MLP) that operates with less than 1\% of the computational footprint of SOTA BC \cite{zarrar2024tinylidarnet} and model-based DRL \cite{brunnbauer2022latent}.

\item Neural Network Interpretation and System Identification: Analysis of the hidden network layers to establish vehicle dynamics models from a functional bifurcation where the first layer extracts discrete spatial states from the high-dimensional observation manifold that are subsequently utilized by the second layer to compute nonlinear vehicle dynamics considerate continuous controls through inhibitory and excitatory pathways. The policy implicitly learns a quasi empirical Pacejka tire model to apply steering commands that push the vehicle into nonlinear saturation by maintaining the slip angle at the boundary of the tire friction circle.

\end{itemize}

\section{Related Work}

Autonomous racing is a subject of extensive research investigations as it serves as a rigorous testbed for autonomy algorithms, at the limits of vehicle capabilities. Existing approaches can be broadly categorized into hierarchical motion planning of precomputed reference paths via classical, ML or hybrid techniques, which rely on prebuilt maps \cite{ghignone2025rlpp, liniger2015optimization}, and end-to-end kinodynamic ML models which simultaneously solve path and motion planning from onboard sensor observations with generalization to new track configurations \cite{zarrar2024tinylidarnet, evans2023high}. 

The majority of racing research is dedicated to solving the trajectory tracking problem, where a global map and a precomputed optimal reference path \cite{heilmeier2020minimum, sivashangaran2022nonlinear} are utilized. This is the predominant method in full-size autonomous racing, which is at its infancy, hence requires tried-and-tested safety critical techniques. In full-scale competitions such as the Indy Autonomous Challenge (IAC) and Formula Student Driverless, vehicles commonly utilize hierarchical, modular perception and planning stacks \cite{betz2023tum, Pinho2023LearningBasedMP}. These frameworks utilize optimization algorithms such as Model Predictive Control (MPC) and derivatives that comprise Learning Model Predictive Control (LMPC) \cite{Joa2024PiecewiseAR, costa2023online}, and Nonlinear Model Predictive Control (NMPC) \cite{vazquez2020optimization, cataffo2022nonlinear}, to maintain vehicles within track boundaries while maximizing tire forces. Although mathematically optimal in controlled settings, these systems are intrinsically brittle, with success predicated on a high-frequency continuous localization pipeline and the availability of detailed maps. Consequently, these do not adapt to navigate OOD dynamic, unmapped environments.

Driven by the advancement of superhuman Artificial Intelligence (AI) in driving video games \cite{fuchs2021super} and physical drone racing \cite{kaufmann2023champion}, several works investigate DRL, by adapting it as a motion planner to learn to track a time-optimal reference trajectory. These match MPC's performance with precise external motion capture systems and finetuning on physical hardware, following simulation pretraining \cite{chisari2021learning}. Motivated by solving the more complex fully-onboard robotics problem, hybrid residual architectures \cite{ghignone2025rlpp} represent the SOTA in sim-to-real reference tracking DRL, but do not yet match MPC, with noisy localization and state estimation. These frameworks mitigate the high-frequency slaloming of pure neural controllers by restricting the DRL policy to output minor corrections atop a classical tracking controller, such as Pure Pursuit (PP). Although trajectory tracking DRL agents achieve near optimal performance in the training track configuration, these do not generalize OOD as well as end-to-end agents that use onboard depth measurements and function independent of a precomputed trajectory and prebuilt map \cite{evans2023comparing}. The parameterized kinodynamics remain tethered to the mathematical assumptions of the baseline tracker, rendering robust mapless OOD generalization ineffective without track-specific retraining. Consequently, full planning agents that use a prebuilt map, but utilize track centerline points and depth measurements instead of a reference trajectory, generalize OOD better than trajectory tracking agents.

Achieving true mapless navigation, where the policy infers feasible trajectories and nonlinear dynamics solely from instantaneous sensor data, remains a grand challenge, that is only partially solved in simulation. The methods fail upon physical deployment primarily due to the reliance on low-fidelity simulation tools \cite{o2020f1tenth}. The F1TENTH organization, recently renamed RoboRacer, has since introduced an updated 3D simulator based on the Unity engine \cite{AutoDRIVE-Simulator-2021}. Although its rendering pipeline is exceptional for bridging the visual sim-to-real gap in camera-based applications, Unity was fundamentally built as a game engine, hence its underlying physics systems are optimized for visual plausibility and real-time gaming performance rather than analytical accuracy. Game physics engines utilize mathematical approximations that frequently cause control models to fail when deployed to real hardware, due to a lack of continuous-time contact mechanics and actuator dynamics. Furthermore, running parallelized DRL environments within a game engine is computationally less efficient than in GPU-optimized robotics simulators. Consequently, SOTA sim-to-real DRL robotics policies for tasks ranging from humanoid whole body control, dexterous manipulation and agile locomotion \cite{yang2025omniretarget, liao2025beyondmimic, he2025viral, zhang2026ame} utilize dedicated physics engines such as Bullet, MuJoCo, and Isaac Sim, developed specifically to close the physics reality gap.

Among the end-to-end DRL methods that generalize to new track layouts, ranked in order of speed in simulation, prior works utilize continuous model-free off-policy algorithms such as Twin-Delayed Deep Deterministic Policy Gradient (TD3) and Soft Actor-Critic (SAC) \cite{haarnoja2018soft} trained with prescriptive reference trajectory aided rewards \cite{evans2023comparing}, discrete action algorithms such as Deep Q-Network (DQN) that output discretized left, right, or slowdown commands \cite{bosello2022train}, and model-based methods such as Dreamer \cite{brunnbauer2022latent} trained in the Bullet physics engine to circumvent sim-to-real inaccuracies unlike the former two which used the F!TENTH simulator. Although Dreamer was the slowest, it was the only method to successfully deploy on non-simplified physical track configurations. However it did not solve high-frequency actuation instability, mirroring the physical deployment of policies trained in the F1TENTH simulator. Moreover, on the simplest simulation track that a model-free off-policy algorithm that Dreamer was compared to, was successful, it outperformed the model-based DRL policy. Furthermore, on-policy algorithms such as PPO, which is the standard for performance in AI applications such as Large Language Model (LLM) post-training and humanoid whole body control, are reported to fail. This is predominantly due to geometric tracking rewards that explicitly penalize the mathematical deviation from precomputed steering actions or spatial references, negative collision penalties that enforce conservative behavior, especially pronounced in on-policy algorithms that do not utilize a replay buffer of experience rollouts to sample-efficiently converge to the optimal reference in the reward formulation, immediate accumulation of failed samples in the rollout at high speeds that provide no actionable learning signal, and insufficient training scale. These formulations converge the neural network into a regime of pure geometric path-following, denying the opportunity to encode rigid-body dynamics, which PPO excels in for complex robotics tasks such as agile humanoid whole body control \cite{liao2025beyondmimic}. 

To attain mapless end-to-end control bypassing the sim-to-real gap, recent approaches utilize BC \cite{zarrar2024tinylidarnet}. ML architectures utilizing Convolutional Neural Networks (CNNs) trained directly on real world human demonstrations, successfully deploy on hardware without prior maps. While these BC models demonstrate sample efficiency, requiring as few as 10,478 training samples compared to the millions required by DRL, the reliance on human sourced data enforces a performance threshold. As these networks mimic the operator, the theoretical maximum performance is intrinsically bottlenecked by human reaction times, preventing dynamics encoding at the limits of the lateral friction circle. Thus, in simulated environments, BC is a fraction of DRL's performance, however, since DRL requires a significantly larger scale of training data and collisions at the limit, for parameterization of the best policy that are infeasible without simulation training, BC represents the SOTA for end-to-end ML, due to the sim-to-real inaccuracies.

\section{Physics-Informed Reinforcement Learning of Spatial Density Velocity Potentials} \label{se:method}

Fundamental to the versatility and zero-shot sim-to-real attributes of the policy are the unified parameterization of spectral spatial density to vehicle dynamics with a physics-informed physics engine exploit-aware reward structure. By interpreting instantaneous range observations as a spectral signal rather than geometric occupancy, we project the unbounded geometric potential directly onto the feasible vehicle dynamics manifold, directing the ANN, a 2-hidden layer MLP with 64 nodes in each layer and $tanh$ activation, to shift its optimization objective from prescriptive geometric mimicry to autonomous kinodynamic exploration. Consequently, the agent learns an implicit inverse dynamics model, encoding track features in the first hidden layer, and the nonlinear correlations between spatial variance and permissible actuation in the second, to maximize momentum within the limits of the friction circle. To facilitate this encoding without policy collapse, the framework utilizes a low-speed training environment that implicitly serves as a kinodynamic scaffold, providing the stabilizing benefits of curriculum training without intermediate scheduling. We trained the policy in a Bullet physics engine based simulator \cite{sivashangaran2023autovrl}, and physically validated it with a proportionally scaled car \cite{sivashangaran2023xtenth}.

\subsection{Spatial Spectral Dynamics Parameterization}

The explicit absence of trajectory or spatial references in the reward formulation fundamentally alters ANN parameterization. To evaluate the interaction with the spectral observation space, we parameterized this spatial-kinodynamic inference utilizing two distinct continuous-control model-free actor-critic algorithms.

\textbf{PPO (On-Policy).} The policy is optimized with the objective function $L_{PPO}$ defined as follows, where $t$ is a discrete time step, $\hat{\mathbb{E}}_{t}$ denotes the empirical average over a batch of samples, $r_{t}$ is the probability ratio between the new and old policies $\frac{\pi(a_{t}|s_{t})}{\pi_{old}(a_{t}|s_{t})}$, $a_{t}$ is the executed action at observed state $s_{t}$, $\hat{A}_{t}$ is the Generalized Advantage Estimation (GAE) computed over $t_{GAE}$ time steps with discount factor $\gamma$, and $\epsilon$ is a clipping hyperparameter that constrains the update magnitude to ensure stable learning.

\begin{align}
L_{PPO}=\hat{\mathbb{E}}_{t}[min(r_{t}\hat{A}_{t},clip(r_{t},1-\epsilon,1+\epsilon)\hat{A}_{t})] \label{eqn1}
\end{align}

The policy update is driven by the GAE, while the critic network $V(s_t)$ predicts the velocity potential field. By operating within the curriculum kinodynamic scaffold, the agent experiences the survival horizon necessary to discover and map features without immediate episode truncation. Furthermore, instead of trivially mapping high-frequency spatial constraints to low values, it learns to assign high values to specific high-frequency spectral features, such as corner apexes, that correspond to local maxima in the permissible dynamics manifold. This distinction differentiates between obstacles to be avoided and geometric limits to be exploited for momentum conservation. The network thus creates a deterministic gradient of velocity potential $\frac{\partial\pi}{\partial\mathcal{s}}$ that guides the trajectory to maximize these constraints. As PPO’s on-policy optimization does not prioritize entropy maximization, the network converges to a low-entropy policy that precisely rides the edge of the friction circle, fully utilizing the track limits.

\textbf{SAC (Off-Policy).} An entropy objective is maximized, to balance exploration and exploitation using a stochastic actor updated by minimizing the objective function $L_{SAC}$, where $M$ is the mini-batch size, $Q_{tk}$ are multiple $k$ target critics, and $h$ is the entropy temperature parameter.

\begin{equation}
L_{SAC} = \frac{1}{M} \sum_{t=1}^{M} \left( h \ln \pi(a_{t}|s_{t}) - \min_{k} Q_{tk}(s_{t}, a_{t}) \right) \label{eqn2}
\end{equation}

The equivalent to the advantage function is derived from the Soft Q-function and the entropy-augmented target value $y_{t}$, calculated as the sum of the minimum discounted future reward $R_t$, discounted by $\gamma$, and the weighted entropy.

\begin{equation}
y_{t} = R_{t} + \gamma \left( \min_{k} Q_{tk}(s_{t+1}, a_{t+1}) - h \ln \pi(a_{t+1}|s_{t+1}) \right) \label{eqn3}
\end{equation}

The entropy regularization $h \ln \pi$ acts as a repulsive field against the high-frequency boundaries of the observation space, and prioritizes a broader distribution of actions, thus it intrinsically maps high spatial frequency to uncertainty, without reference trajectory aided rewards, yielding a conservative safety buffer that does not maximize track limits.  

\subsection{Observation Space}

Depth measurements obtainable from LiDAR or camera RGB-D point clouds comprise the observation space. The number of rays $n_{r}$ has a bearing on successful policy learning. Prior model-free algorithms trained in the F1TENTH simulator use 20 rays, and report that an increase does not improve performance \cite{evans2023comparing, bosello2022train}, whereas 1080 rays were used in the Bullet physics engine, which did not yield adequate model-free policies, necessitating model-based DRL \cite{brunnbauer2022latent}. Utilizing Bullet, our observation space $\mathcal{O}$ comprises 170 rays spanning a $120^{\circ}$ frontal Field of View (FoV), defined as follows, where $d_{i}$ is the depth measurement at each angle increment. 

\begin{equation}
\mathcal{O} = [d_i]^{1 \times 170}
\end{equation}

$\mathcal{O}$ is interpreted not merely as geometric occupancy, but as a spectral signal, that is, high-velocity corridors are characterized by low spectral density where the spatial frequency variance $|\nabla \mathcal{O}| \approx 0$, while constraints manifest as high-frequency discontinuities. The neural policy maps these spectral densities directly to permissible velocity potentials using the non-geometric mimicry, physics-informed reward. 20 rays were insufficient, whereas rays in excess of 300 did not improve performance, hence we chose 170 empirically for the Pareto optimization of training time, computation efficiency and performance. 

\begin{figure}[!h]
    \centering
    \includegraphics[width = 0.7\columnwidth]
    {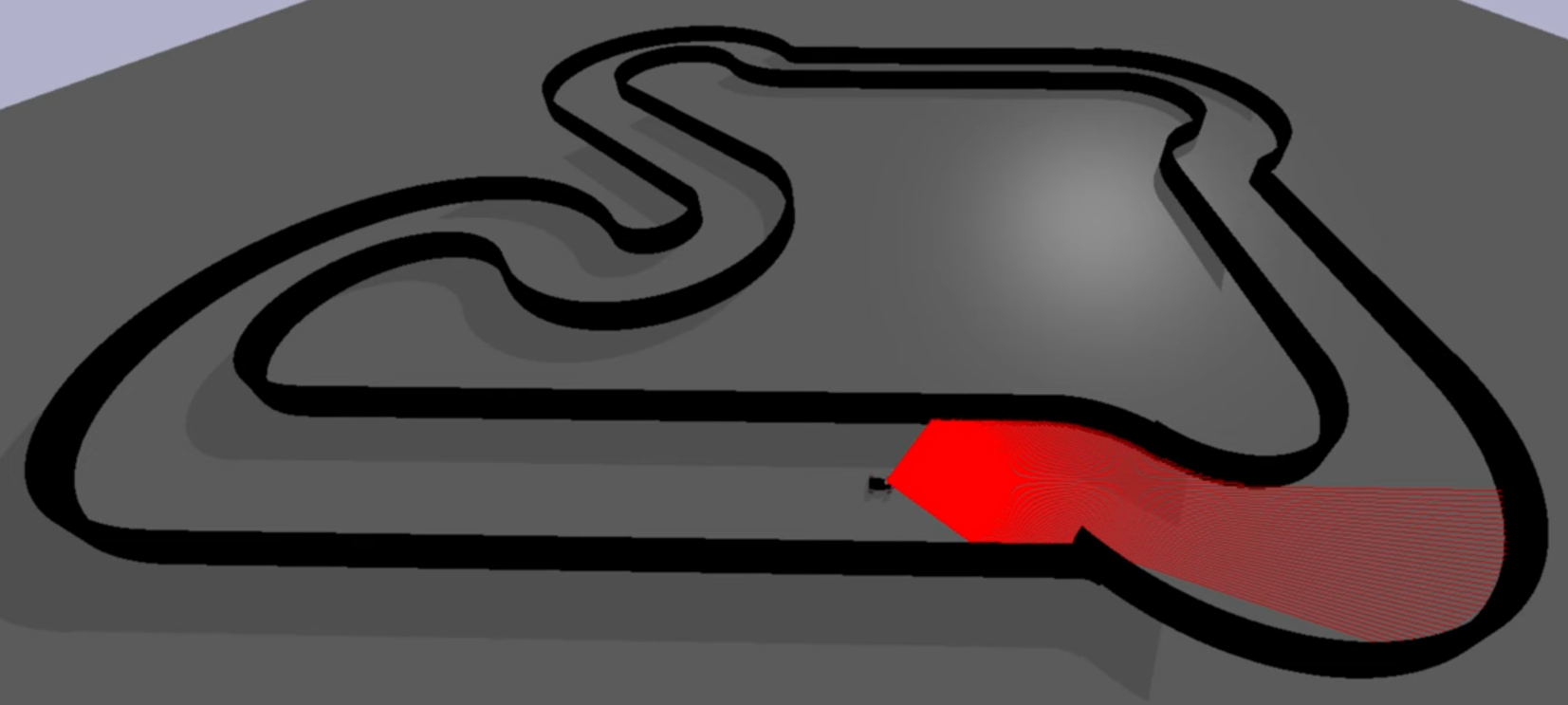}
    \caption{Track boundary detection with simulated spectral depth rays.} \label{racetrack_sim_raycast}
\end{figure}

\textbf{Depth Ray Modeling in Simulation.} The start $(x_{RF}, \allowbreak y_{RF}, z_{RF})$ and end $(x_{RT}, y_{RT}, z_{RT})$ coordinates of depth rays were used as inputs to the physics engine's ray tracing function to obtain the object hit position of each ray $(x_{HP}, y_{HP})$ using which the Euclidean distance from the depth sensor's position to detected obstacles $d_{o}$ was determined as follows. Figure \ref{racetrack_sim_raycast} illustrates the rays detecting simulated track boundaries.

\begin{equation} \label{}
d_o = \sqrt{(x_{HP} - x_{RF})^{2} + (y_{HP} - y_{RF})^{2}}
\end{equation}

$(x_{RF}, y_{RF}, z_{RF})$ were computed as follows where $x_{L,A}$, $y_{L,A}$ and $z_{L,A}$ are coordinates of the depth sensor with respect to the body frame of the Autonomous Ground Vehicle (AGV), $x_{A}$, $y_{A}$ and $z_{A}$ represent the position of the AGV in the global coordinate frame, and $\psi_{A}$ is the AGV's yaw.

\begin{equation} \label{}
x_{RF} = x_{L,A}cos(\psi_{A}) + x_{A}
\end{equation}

\begin{equation} \label{}
y_{RF} = y_{L,A}sin(\psi_{A}) + y_{A}
\end{equation}

\begin{equation} \label{}
z_{RF} = z_{L,A} + z_{A}
\end{equation}

$(x_{RT}, y_{RT}, z_{RT})$ of $n_{r}$ rays each with a maximum length of $d_{max}$ were computed as follows where $(\frac{\pi}{4} - \psi_{A})$ is the LiDAR's orientation, and $\theta(\frac{1:n}{n})$ segments the FoV $\theta$ into $n_{r}$ equally spaced rays.

\begin{equation} \label{}
x_{RT} = d_{max}sin\left((\frac{\pi}{4} - \psi_{A}) + \theta(\frac{1:n_{r}}{n_{r}})\right)
\end{equation}

\begin{equation} \label{}
y_{RT} = d_{max}cos\left((\frac{\pi}{4} - \psi_{A}) + \theta(\frac{1:n_{r}}{n_{r}})\right)
\end{equation}

\begin{equation} \label{}
z_{RT} = z_{L,A}
\end{equation}

\textbf{Physical Depth Sensor Pre-Processing.} Depth measurements for $\mathcal{O}$ can be inferred from LiDAR or cameras such as stereo and infrared. We tested two $360^{\circ}$ 2D LiDARs, the YDLIDAR G2 and RPLIDAR S2, and the Intel RealSense D435i infrared camera. Each was pre-processed to filter outliers, and spatially shaped to fit $\mathcal{O}$. Among these, RPLIDAR S2 yielded the best performance as it provides seven times the data density of YDLIDAR G2, and is faster with a wider horizontal FoV than the Intel RealSense D435i. 

The raw $360^{\circ}$ data was processed as follows. Once null values and other outliers are filtered, these entries are estimated by interpolating from the nearest non-zero observations that precede and follow. Next, the subset of observations within a $120^{\circ}$ FoV directly in front of the AGV are extracted symmetrically, within the range of $-60^{\circ}$ to $+60^{\circ}$ degrees for a default $0^{\circ}$ forward direction. This high resolution data is uniformly downsampled to $n_{r}$.

\begin{algorithm}
\caption{Depth Sensor Processing}

\begin{algorithmic}[1]
\State \textbf{Input:} $O_{s} \gets$ Raw RPLiDAR S2 observations (360° FoV)
\State \textbf{Output:} $O \gets$ Processed observations

\State \textbf{Step 1: Outlier Handling}
\For{$i \gets 1$ \textbf{to} $\text{length}(O_{s})$}
    \If{$O_{s}[i]$ is missing}
        \State $O_{s}[i] \gets 0$
    \EndIf
\EndFor

\State \textbf{Step 2: Interpolation of Missing Values}
\For{$i \gets 1$ \textbf{to} $\text{length}(O_{s})$}
    \If{$O_{s}[i] = 0$}
        \State $prev \gets \text{max}(O_{s}[1:i-1] \text{ where } O_{s} \neq 0)$
        \State $next \gets \text{min}(O_{s}[i+1:\text{end}] \text{ where } O_{s} \neq 0)$
        \State $O_{s}[i] \gets \frac{prev + next}{2}$
    \EndIf
\EndFor

\State \textbf{Step 3: Field of View Selection}
\State $O_{s, fov} \gets O_{s}[\theta_{start} : \theta_{end}]$ where $\theta_{start} = -60^\circ$, $\theta_{end} = +60^\circ$

\State \textbf{Step 4: Downsampling}
\State $n_{r} \gets 170$
\State $step_{size} \gets \frac{\text{length}(O_{s, fov})}{n_{r}}$
\For{$i \gets 1$ \textbf{to} $n_{r}$}
    \State $idx \gets \text{round}(i \times step_{size})$
    \State $O[i] \gets O_{s, fov}[idx]$
\EndFor

\State \textbf{Return:} $O$
\end{algorithmic}
\end{algorithm}

\subsection{Action Space}

The policy's actions comprise the controls for a car, throttle $T \in [0\; 1]$ and steering angle $\delta \in [\delta_{min}  \delta_{max}]$. Notably, braking is omitted to mimic the controls of the scaled car used for validation in contemporary research.

The normalized continuous action vector $a = (a_T, a_\delta);\; \allowbreak a_{T}, a_{\delta} \in [-1, 1]$ computed by the policy was converted to $T$ and $\delta$ as follows, where $\delta_{min}$ and $\delta_{max}$ are $-0.36^{c}$ and $0.36^{c}$.

\begin{equation} \label{}
T = min(max(a_{T}, 0), 1)
\end{equation}

\begin{equation} \label{}
\delta = max(min(a_{\delta}, \delta_{max}), \delta_{min})
\end{equation}

\textbf{Friction Modeling in Simulation.} To encode nonlinear dynamics and traction management, necessary for time-optimal racing, we used the following model to compute the target motor velocity from $T$. The use of $T$ with this friction model, instead of a simplified motor force command \cite{brunnbauer2022latent} grounds the learned policy in realistic vehicle dynamics, requiring the agent to navigate the nonlinear boundaries of tire adhesion rather than relying on a frictionless acceleration proxy.

The resistive force $f$ was computed as follows where $C_{d}$ and $C_{r}$ are the drag and rolling resistance constants, and $v_{m,t-1}$ is the target motor velocity in the prior time step.

\begin{equation} \label{}
f = v_{m,t-1}(v_{m,t-1}C_{d} + C_{r})
\end{equation}

The motor acceleration for the current time step $\dot{v}_{m,t}$ was computed from the throttle input $T$, throttle constant $C_T$ and resistive force $f$ as follows.

\begin{equation} \label{}
\dot{v}_{m,t} = C_{T}T - f
\end{equation}

The target motor velocity for the current time step $v_{m,t}$ to be applied, was computed as follows, where $t_{s}$ is the sampling time.

\begin{equation} \label{}
v_{m,t} = v_{m,t-1} + t_{s}\dot{v_{m,t}}
\end{equation}

Constants $C_T$, $C_{d}$ and $C_{r}$ are variable parameters set to 20, 0.01 and 0.2, used to tune rate of acceleration and resistance. $C_T$ was empirically determined for training convergence. For post-training evaluation at the physical car's limits, we increased $C_T$ to 120.

$v_{m,t}$ was applied to the simulated AGV via the PyBullet motor control function with a motor force of $1.2 N$, and is proportional to the AGV's velocity $v$. This nonlinear coupling between $T$ and $v$ is critical for the reward formulation.

\subsection{Physics-Informed Simulator-Exploit Aware Reward}

Reward objectives are minimally prescriptive and comprise two explicit formulas and an implicit component. In contrast to prior works, we do not use precomputed reference trajectories or track progress, and instead derive a sparse formulation that couples the reward directly to the vehicle's equilibrium dynamics. This necessitates the elimination of an explicit penalty after episode termination upon contact with the track boundary, without which the ANN parameterizes conservative boundary avoidance, without geometric targets. Instead, collisions are penalized by an implicit truncation of the value function, which can be visualized as the lost opportunity cost. Furthermore, high-frequency actuation instability is an inhibitor of sim-to-real transfer in continuous control autonomous racing, diagnosed as a classical time-optimal control artifact \cite{brunnbauer2022latent}. 
To resolve this sim-to-real gap, we identify the instantaneous transitions between actuator limits, perceived as a mathematically costless kinodynamic action due to physics engine inaccuracies, and apply an oscillation penalty to prune the non-physical physics engine exploit.  

\textbf{Self-Supervised Velocity Potential.} The primary learning signal is a physics-informed sparse throttle reward, $R_{T} = T^2$ that organically solves the optimization problem to maximize the velocity potential $\Phi(v) \propto v^4$ without prescriptive geometric mimicry inferred by evaluating the steady-state limit at $\dot{v} = 0$, $T_{ss}$ as follows.

\begin{equation}
T_{ss} = \frac{C_{d} v^2 + C_{r} v}{C_T} \implies R_{T} \propto v^4
\end{equation}

This projects the unbounded geometric potential onto the feasible vehicle dynamics manifold, encouraging the agent to seek a maximum $\Phi(v)$ action $a_{t}$ subject to the physical constraint where the centripetal force required for the turn, defined by vehicle mass $m$, velocity $v$, and track curvature $\kappa$ estimated from observations, does not exceed the available tire friction force $\mu F_{z}$ where $\mu$ is the coefficient of friction and $F_{z}$ is the normal force.

\begin{equation}
a_t \text{ s.t. }\frac{mv^{2}}{\kappa}\le\mu F_{z}
\end{equation}

A scalor multiplier of five, determined empirically, is applied to $R_{T}$ to balance the weight of the auxiliary penalty.   

\textbf{Implicit Collision Value Truncation.} Explicit collision penalties mathematically prioritize crash minimization over optimal speed, injecting variance-induced conservatism that manifests as timid driving and excessive wall avoidance, particularly pronounced with $R_{T}$. To discourage collisions, but maximize the friction circle, we treat collisions as an implicit truncation of the value horizon $V(\mathcal{O}_t)$. Upon episode termination, $V(\mathcal{O}_t)=0$ instead of $V(\mathcal{O}_t)\approx\sum_{k=0}^{\infty}\gamma^kT_{}^2$, transforming the penalty from an arbitrary negative scalar into a dynamic lost opportunity cost. Thus, the mathematical cost of a collision converges to approximately $\frac{1}{1-\gamma}$. Consequently, the agent is incentivized to avoid track boundaries not out of programmed repulsion, but to continue accumulating maximum positive velocity rewards.

\textbf{Simulator Exploit Pruning.} To prevent steering instability that limits real world policy transfer, we introduce a targeted kinodynamic constraint. In prior work, this phenomenon was interpreted as an artifact of classical time-optimal control and classified as bang-bang steering \cite{brunnbauer2022latent}, thus was attempted to be resolved by continuous action smoothing with $L_2$ regularization penalties to both the magnitude and rate of steering.

Although this improved sim-to-real transfer, it did not mitigate it. Our analysis indicates that this behavior is a direct exploitation of discrete-time simulation mechanics which mathematically permit instantaneous transitions between actuator limits due to inaccurately modeled physical factors such as servomotor delay, chassis load transfer and tire scrubbing resistance that would naturally destabilize the car upon full-lock-to-full-lock steering reversal.

Consequently, the agent learns to perceive high-frequency oscillation as a mathematically costless action. Evidence from both the prior correction architecture and our framework corroborate that introducing a specific oscillation penalty yields no alteration to simulation performance or convergence efficiency, but exclusively improves sim-to-real transfer viability. If slaloming was a requisite for time-optimal control, mathematically penalizing it would inherently alter the agent's lap times within the simulation. As the theoretical optimum in simulation remains unchanged but real world performance improves, we posit bang-bang steering is analytically a simulator loophole rather than a theoretical optimization flaw. 

To resolve this sim-to-real gap we use a precise oscillation penalty $R_{StR}$ to prune the identified non-physical physics engine exploit as follows, where $a_{\delta,t}$ and $a_{\delta,t-1}$ are the normalized steering angles at the current and previous time steps.

\begin{equation}
R_{StR} = -2.0 \quad \text{if} \quad a_{\delta, t} \cdot a_{\delta, t-1} = -1
\end{equation}

This formulation explicitly isolates the specific min-max full-lock-to-full-lock steering reversal. By pruning this exact non-physical physics engine exploit, rather than broadly penalizing the steering rate, the policy maintains optimal cornering performance while completely eliminating slaloming on physical hardware.

\textbf{Reward Ablations.} To validate the reward formulation $R = 5R_{T} + R_{StR}$, ablation studies were conducted, comparing reward elements for convergence speed and performance. The ablations are listed in Table \ref{reward_ablation} where $d_{min}$ is the minimum depth measurement in $\mathcal{O}$ and $d_{coll}$ is the distance at which collision occurs. $R_{StR}$ was added to each ablation formula.

\begin{table}[h]
    \renewcommand{\arraystretch}{1.2}
    \centering
    \caption{Reward Ablation Experiments}
    \label{reward_ablation}
    \resizebox{\columnwidth}{!}{
        \begin{tabular}[t]{|c|c|c|}
            \hline
            \textbf{Ablation} & \textbf{Description}& \textbf{Formula} \\
            \hline
            $R$ & Squared throttle, $T^{2}$ & $5T^{2}$ \\
            \hline
            $R_{ab1}$ & Unsquared throttle, $T$ & $5T$ \\
            \hline
            $R_{ab2}$ & Collision penalty, $-1$ & $5T^2$ \& -1 \text{if } $d_{min} < d_{coll}$ \\
            \hline
        \end{tabular}
        }
\end{table}

\subsection{Simulation Training \& Physical Deployment}

An Intel Core i9 13900KF CPU and an NVIDIA GeForce RTX 4090 GPU were used for training, conducted over 20,000,000 time steps, encompassing 15,747 collisions, in 48 wall clock hours. This scale is infeasible in the real world. By allowing catastrophic failures in simulation, the agent fully explores the state-action space boundaries, encoding the complex correlation between spatial features and vehicle dynamics. The training hyperparameters are summarized in Table \ref{parameters}.

\begin{table}[ht]
    \renewcommand{\arraystretch}{1.2}
    \centering
    \caption{Training Hyperparameters}
    \label{parameters}
        \begin{tabular}[t]{|c|c|}
            \hline
            \textbf{Hyperparameter} & \textbf{Value} \\
            \hline
            Discount Factor ($\gamma$)  & 0.99 \\
            \hline
            Learning Rate & 0.0003 \\
            \hline
            Rollout Buffer Size & 2048 \\
            \hline
            Batch Size & 64 \\
            \hline
            Number of Epochs & 10 \\
            \hline
        \end{tabular}
\end{table}

The comparison policies which comprise SAC and the reward ablations were trained with the same parameters for up to 50,000,000 time steps, with model checkpoints saved every 5,000,000. The best performer in each was used for validation.

\begin{figure}[htbp]
\centering
\begin{minipage}{0.2\textwidth}
\includegraphics[width=0.9\textwidth]{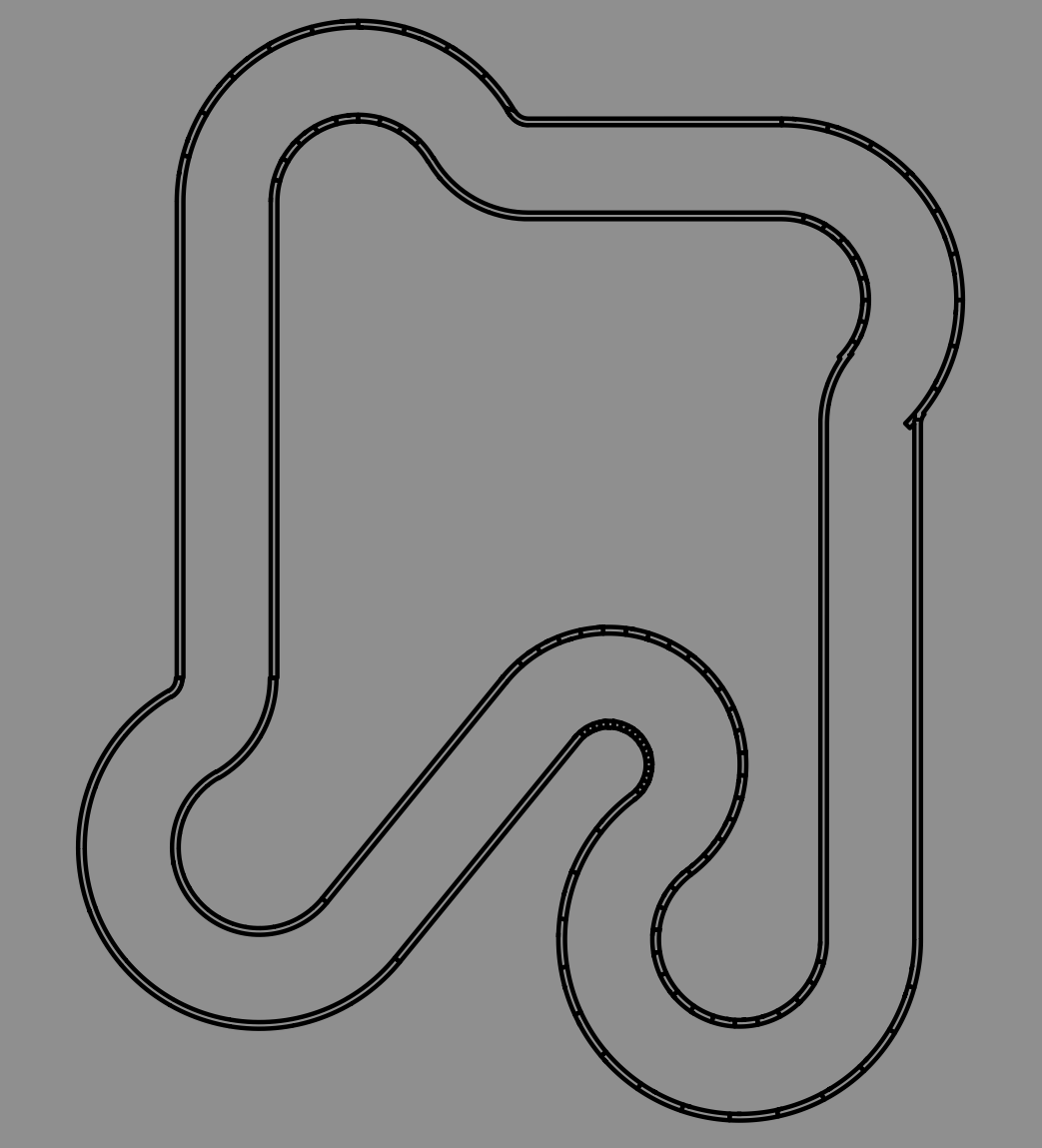}
\subcaption{} \label{racetrack1}
\end{minipage}%
\\
\begin{minipage}{0.2\textwidth}
\includegraphics[width=0.9\textwidth]{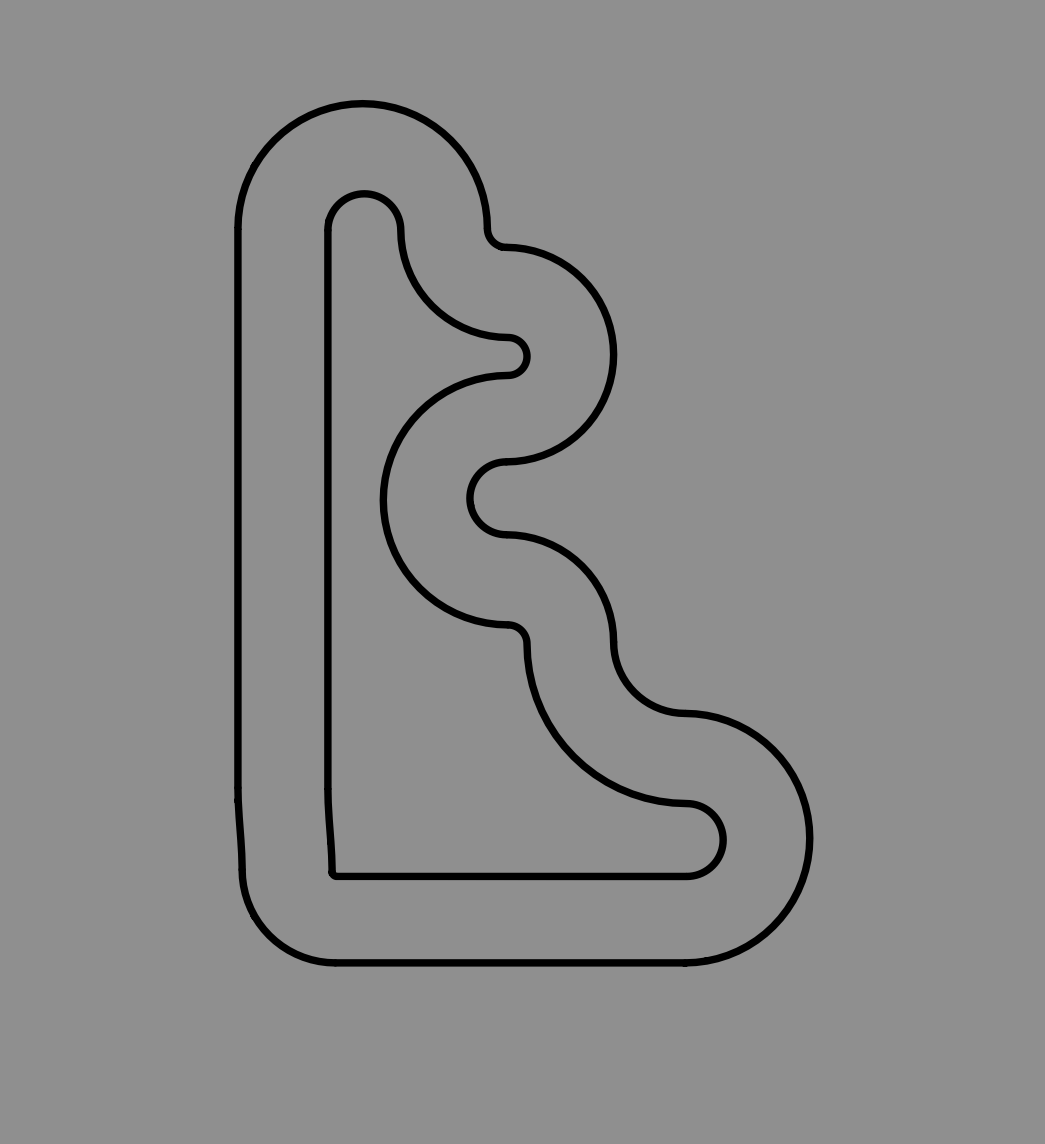}
\subcaption{} \label{racetrack2}
\end{minipage}
\begin{minipage}{0.2\textwidth}
\includegraphics[width=0.9\textwidth]{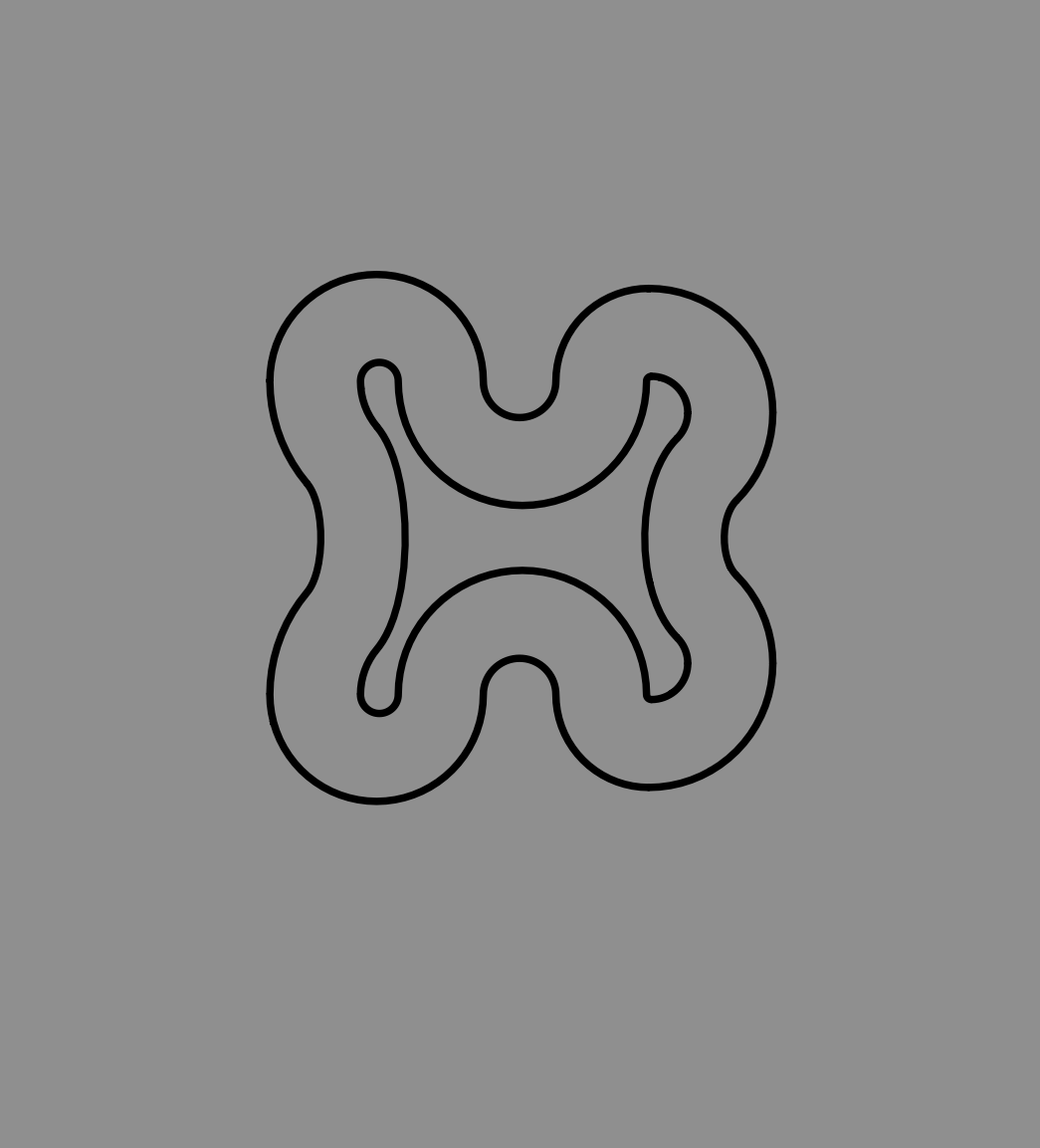}
\subcaption{} \label{racetrack3}
\end{minipage}%
\caption{Simulated track layouts. (a) Training Track. (b) OOD Track 1. (c) OOD Track 2.} \label{racetracks_sim_layout}
\end{figure}

\begin{figure}[htbp]
\centering
\begin{minipage}{0.2\textwidth}
\includegraphics[width=0.9\textwidth]{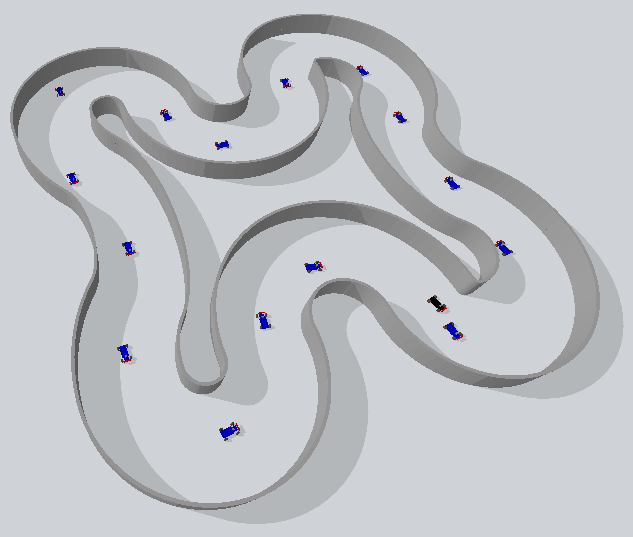}
\subcaption{} \label{racetrack_ot}
\end{minipage}%
\begin{minipage}{0.2\textwidth}
\includegraphics[width=0.9\textwidth]{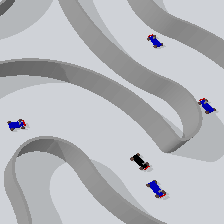}
\subcaption{} \label{racetrack_ot_zoom}
\end{minipage}
\caption{Multi-agent overtake environment in OOD Track 2. (a) Obstacle cars lapping with the pretained DRL policy. (b) Ego car trained to overtake in black.} \label{racetrack_ot_layout}
\end{figure}

\textbf{Simulation Infrastructure.} Due to the lack of simulation assets in analytically accurate physics engines, we developed multiple track and multi-agent environments, illustrated in Figures \ref{racetracks_sim_layout} and \ref{racetrack_ot_layout} for DRL in the open-source AutoVRL simulator \cite{sivashangaran2023autovrl}, built on the Bullet physics engine \cite{coumans2021}. The tracks were modeled with Onshape, and imported to Bullet in Unified Robot Description Format (URDF). This necessitated segmentation of the straights and corners with explicit specification of concave triangle collision meshes for accurate inner and outer track boundary collision boxes.   

The track in Figure \ref{racetrack1} which occupies 760 $m^2$ with 4 straights and 4 corners in each direction was used for training, and the tracks in Figures \ref{racetrack2}, which occupies 375 $m^2$ with 2 straights, 5 left corners and 3 right corners and \ref{racetrack3} which occupies 210 $m^2$ with no straights and 4 corners in each direction were used for OOD validation. The latter was configured to a multi-agent environment for overtake validation, with 15 obstacle cars and the ego car, as shown in Figure \ref{racetrack_ot}. The obstacle cars, in blue, use the pretrained DRL policy for inference, and the ego car, in black, was trained to overtake with double the $C_T$, as depicted in Figure \ref{racetrack_ot_zoom}. 

\begin{figure}[!h]
    \centering
    \includegraphics[width = 0.6\columnwidth]
    {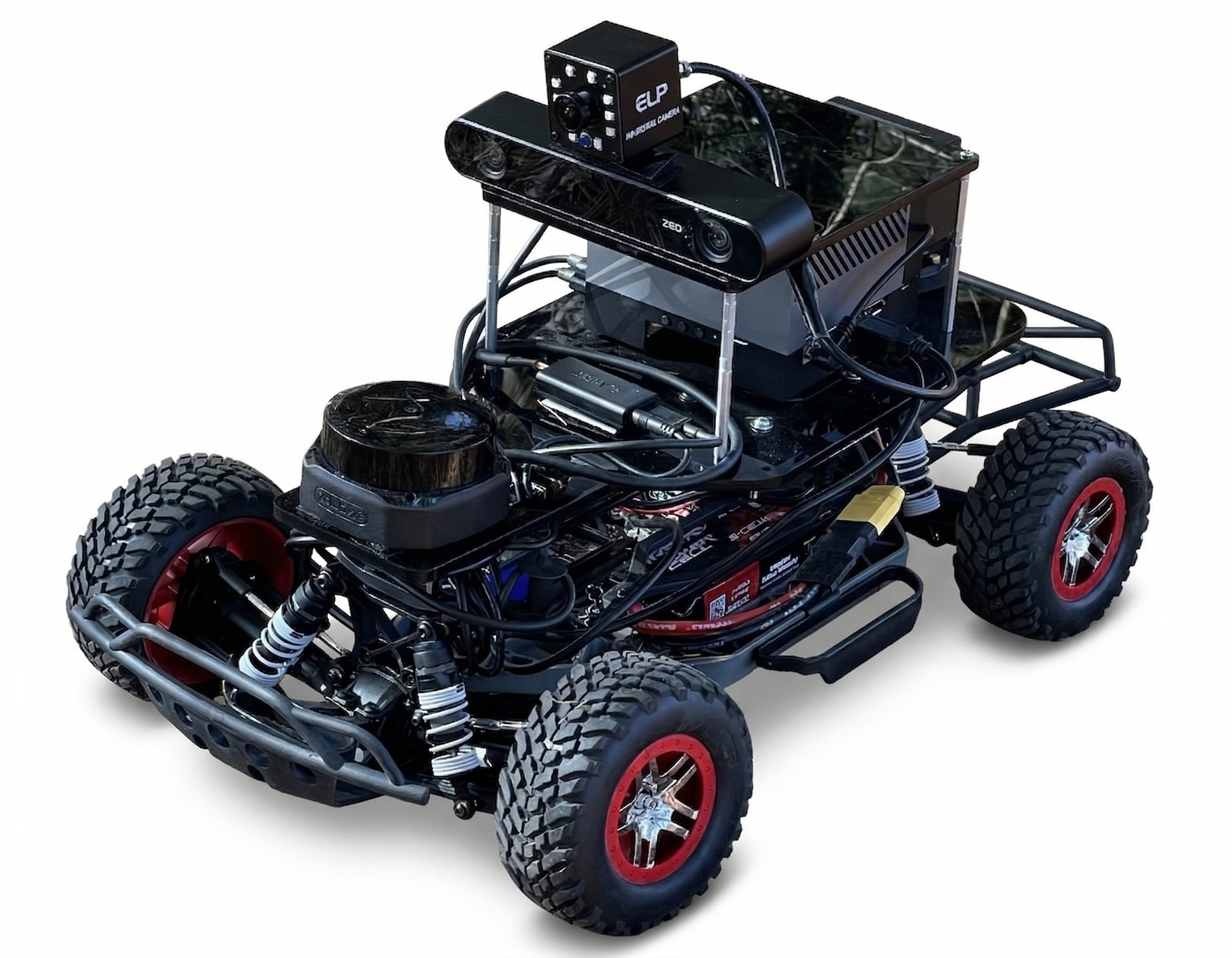}
    \caption{Proportionally 1/10th scaled hardware platform \cite{sivashangaran2023xtenth}.} \label{xtenthcar}
\end{figure}

\textbf{Policy Transfer to Hardware.} The trained model was transferred zero-shot to hardware \cite{sivashangaran2023xtenth}, depicted in Figure \ref{xtenthcar}, by replacing the physics engine's simulated observations and actions with those processed in the scaled car, interfaced with the Robot Operating System (ROS). The physical test tracks occupied 25 $m^2$ comprising 3 layouts each with a different combination of straights and corner directions. These configurations require throttle modulation to minimize lap time, thus to validate smooth hardware transfer at high speeds, a $90^{\circ}$ corner, an angle not in the training track, was utilized for controlled tests at constant 1, 3 and 5 $m/s$.

\section{Neural Network Interpretation}

Racing is a singular optimization problem where at the limit of handling, the vehicle's trajectory is bound by the holonomic constraint of the friction circle, where the lateral force cannot exceed the tire friction force, $F_{lat} \le \mu F_z$, thus the optimal velocity $v_{opt} = \sqrt{\mu g R(\rho)}$ is encoded in the curvature of the wall currently perceived $R(\rho)$, where $g$ is the gravitational acceleration, and is independent of a measured velocity, on the manifold of optimality. As such, the policy learns to solve the control problem at every instant mapping geometry to energy, ensuring the vehicle state remains continuously synchronized with the physical limits of the environment. To analyze how the model computes nonlinear kinodynamic controls from instantaneous spatial measurements, without temporal memory, we dissected the internal activations of the ANN and mapped the resulting control policy to establish vehicle dynamics models. 

\textbf{Inter-Layer Correlation Analysis.} To systematically decode the neural architecture, we applied the Pearson product-moment correlation coefficient $r$ defined as follows, across the telemetry sequence in the training track, to cross-correlate the first and second hidden layers to each other and to the output layer. This metric quantifies the linear dependence between the hidden activations and the vehicle's kinodynamic state, to locate localized, specialized linear pathways. Here $n_{s}$ is the total number of discrete time steps in the telemetry sample, $x_t$ and $y_t$ are the instantaneous values of the two variables $X$ and $Y$ from different network layers being compared such as a neuron's activation state in the first layer and the steering command in the output layer, and $\bar{x}$ and $\bar{y}$ are the mean across the dataset.

\begin{equation}
r = \frac{\sum_{t=1}^{n_{s}} (x_t - \bar{x})(y_t - \bar{y})}{\sqrt{\sum_{t=1}^{n_{s}} (x_t - \bar{x})^2} \sqrt{\sum_{t=1}^{n_{s}} (y_t - \bar{y})^2}}
\end{equation}

\textbf{System Identification of Tire Dynamics.} To model tire physics, we reverse-engineered the spatial inputs and corresponding actions into proxies of standard vehicle dynamics parameters, lateral acceleration $a_{lat,p}$ and slip angle $\alpha_{p}$. 

$a_{lat,p}$ was computed using the instantaneous track curvature $\kappa$ derived from the center ray of the spatial scan $d_{ctr}$, the throttle command $a_{T}$ as a proxy for velocity, the sign of the steering command $a_{\delta}$ to infer the vector direction as follows.

\begin{equation}
\kappa = \text{sgn}(a_{\delta}) \cdot \frac{1}{d_{ctr}}
\end{equation}

\begin{equation}
a_{lat, p} = \kappa \cdot a_{T}
\end{equation}

$\alpha_{p}$ is the deviation between the policy's steering command $\delta$ and the kinematic steering angle $\delta_{k}$ required for a rigid body. Using the vehicle's fixed physical wheelbase $l_{wb} = 0.3$\text{m} and Ackermann geometry, this is computed as follows where $\delta_{o}$ is a constant offset in straights estimated to be 0.08 via linear regression on low-velocity samples.

\begin{equation}
\delta_{k} = l_{wb} \cdot \kappa + \delta_{o}
\end{equation}

\begin{equation}
\alpha_p = \delta - \delta_{k}
\end{equation}

To derive the empirical friction curve, we extracted the $95^{th}$ percentile of $a_{lat,p}$ in uniformly segmented $\alpha_p$, and fitted two curves. The first is a linear kinematic model defined as follows where $C_{\alpha}$ is the linear stiffness.

\begin{equation}
a_{lat,p} = C_{\alpha} \cdot \alpha_{p}
\end{equation}

This simple model restricts optimal control solvers to conservative boundary avoidance, hence classical methods utilize the Pacejka tire model, defined as follows where $B$ is the stiffness factor, $C$ is the shape factor and $D$ is the peak adhesion limit.

\begin{equation}
a_{lat,p} = D \cdot \sin(C \cdot \arctan(B \cdot \alpha_{p}))
\end{equation}

To quantitatively assess how accurately the linear kinematic and nonlinear Pacejka models represent the policy's learned driving behavior, we computed the Coefficient of Determination $R_{CoD}^2$ to evaluate the proportion of the variance in the empirical boundary data that is predictable from the theoretical equations as follows, where $n_{t}$ is the number of discrete $\alpha_{p}$ segments, $\bar{a}_{lat}$ is the mean of the empirical lateral acceleration and $\hat{a}_{lat}$ is the theoretical lateral acceleration derived from either the linear kinematic or nonlinear Pacejka formulations.

\begin{equation}
R_{CoD}^2 = 1 - \frac{\sum_{i=1}^{n_{t}} (a_{lat, i} - \hat{a}_{lat, i})^2}{\sum_{i=1}^{n_{t}} (a_{lat, i} - \bar{a}_{lat})^2}
\end{equation}

\section{Results and Discussion} \label{se:results}

This section systematically evaluates the proposed method across simulated and physical domains. First, we analyze the training algorithms and reward ablations by comparing lap times and theoretical racing line deviations within the training and two OOD tracks at the curriculum speed. The highest-performing policy is scaled to the maximum physical velocity and evaluated across all simulated tracks, followed by zero-shot physical transfer validation on a proportionally scaled vehicle. The network's computational complexity and inference efficiency are assessed against comparative baseline architectures to quantify edge hardware deployment viability, and the adaptability of the framework's physics-informed dynamics encoding is demonstrated through overtaking maneuvers in a dynamic multi-agent environment. Finally, an examination of the internal neural mechanisms and system identification detail the implicit parameterization of nonlinear vehicle dynamics.

\subsection{Simulation Training \& OOD Transfer: Training Algorithm, Reward Ablation \& Velocity Scaling}

The performance of policies trained with PPO and SAC are quantified through lap completion times and the deviation from the theoretical minimum curvature racing line computed with \cite{heilmeier2020minimum}, in the training and OOD tracks. The best performing algorithm is utilized for subsequent reward ablation and velocity scaling analyses.

\begin{table*}[!b]
    \renewcommand{\arraystretch}{1.2}
    \centering
    \caption{DRL Training Algorithm Lap Times ($s$)}
    \label{rl_algorithm_lap_times}
    \resizebox{\textwidth}{!}{%
        \begin{tabular}{|c|c|c|c|c|c|c|c|c|c|c|c|c|}
            \hline
            \multirow{2}{*}{\textbf{Algorithm}} & \multicolumn{4}{c|}{\textbf{Training Track}} & \multicolumn{4}{c|}{\textbf{OOD Track 1}} & \multicolumn{4}{c|}{\textbf{OOD Track 2}} \\
            \cline{2-13}
             & \textbf{$t_{min}$} & \textbf{$t_{max}$} & \textbf{$\bar{t}$} & \textbf{$\sigma_{t}$} & \textbf{$t_{min}$} & \textbf{$t_{max}$} & \textbf{$\bar{t}$} & \textbf{$\sigma_{t}$} & \textbf{$t_{min}$} & \textbf{$t_{max}$} & \textbf{$\bar{t}$} & \textbf{$\sigma_{t}$} \\
            \hline
            PPO & \textbf{52.27} & \textbf{52.37} & \textbf{52.33} & \textbf{0.034} & \textbf{33.40} & \textbf{33.63} & \textbf{33.53} & \textbf{0.073} & \textbf{20.03} & \textbf{24.93} & \textbf{23.61} & 1.86 \\ 
            \hline
            SAC & 57.43 & 63.00 & 59.58 & 1.71 & 37.73 & 39.57 & 38.88 & 0.608 & 28.00$^{4\dagger}$ & 29.80$^{4\dagger}$ & 29.01$^{4\dagger}$ & \textbf{0.683$^{4\dagger}$} \\ 
            \hline
        \end{tabular}%
    }
\end{table*}

\begin{figure}
    \centering
    \includegraphics[width = \columnwidth]
    {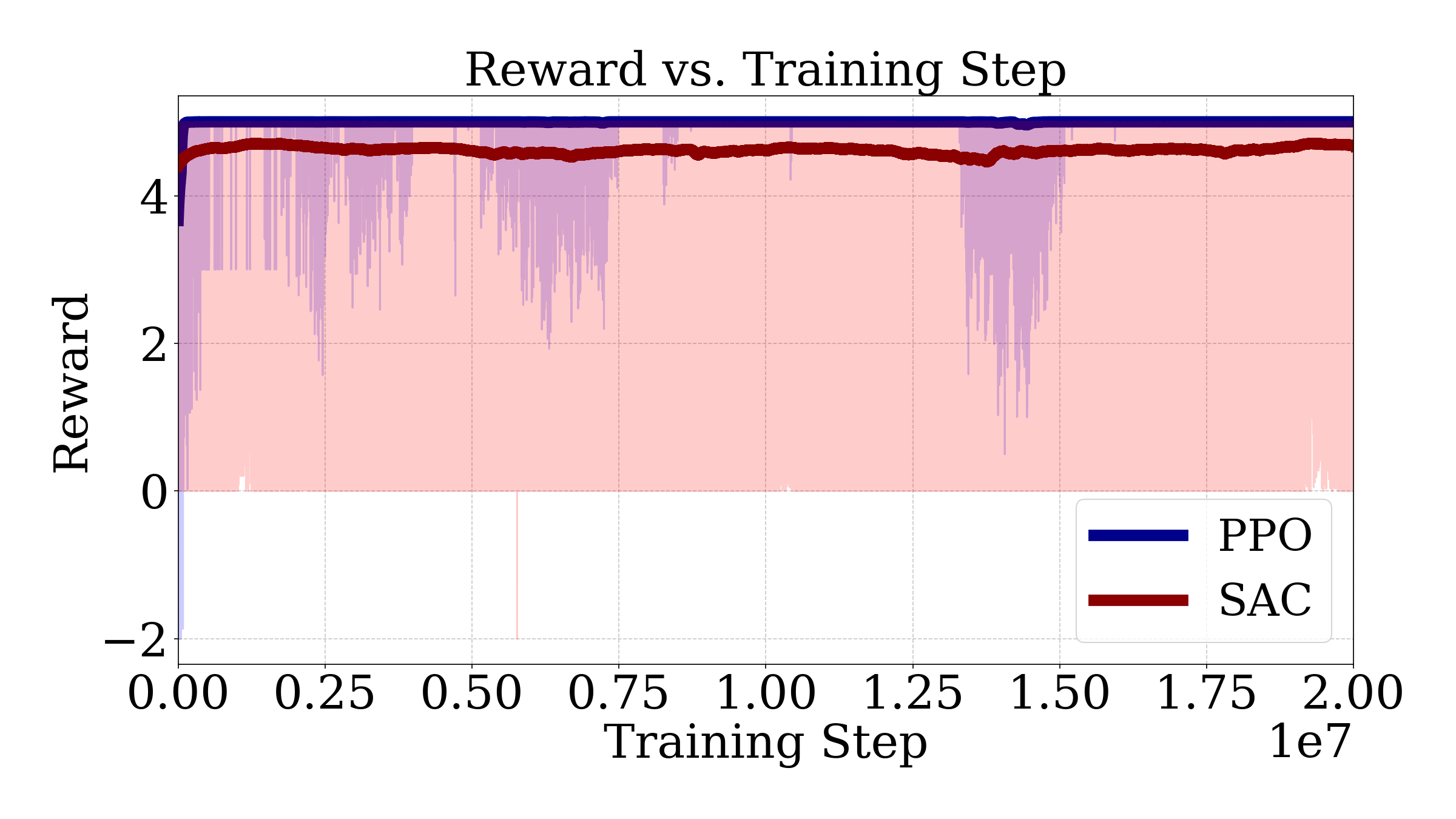}
    \caption{Learning Curve: Moving average of order 100,000 reward at each training step.} \label{plot_learning_curve}
\end{figure}

\textbf{Training Algorithm. } PPO attained a greater reward than SAC, with lower stochasticity. Both learned the best policy at 20,000,000 steps out of the 50,000,000 tested. The learning curves for each are depicted in Figure \ref{plot_learning_curve} which plots the moving average of order 100,000 reward against the training step. 

PPO attained the maximum average reward of 5 at the curriculum speed, successfully applying the throttle at 100\% in every part of the track, by learning the optimal steering to maximize tire friction. Comparatively, SAC achieved 4.68, thus lifted off the throttle to avoid collisions due to suboptimal steering learned to maximize entropy in addition to the reward. Furthermore, PPO corrected the physics engine exploit within the first 50,000 training steps while SAC took just under 6,000,000 to eradicate the penalty. 

The raw data illustrates greater structure in PPO due to the policy's lower entropy. The dips in the reward, that last for several million steps until 15,000,000 training steps indicate the continued improved parameterization past prior convergence to the maximum reward. After 10,000,000 training steps, the PPO policy was able to time-optimally navigate the training track, but collided after 3 laps. Continued training fundamentally altered the latent parameters, as evidenced by the drop around 13,500,000 steps. Drops of lower magnitude persist past 15,000,000 steps, hence the best policy with 5,000,000 training step increments was obtained at 20,000,000.

\begin{table*}[ht]
    \renewcommand{\arraystretch}{1.2}
    \centering
    \caption{DRL Training Algorithm Deviation from Theoretical Baseline ($m$)}
    \label{rl_algorithm_deviation}
    \resizebox{\textwidth}{!}{%
        \begin{tabular}{|c|c|c|c|c|c|c|c|c|c|c|c|c|}
            \hline
            \multirow{2}{*}{\textbf{Algorithm}} & \multicolumn{4}{c|}{\textbf{Training Track}} & \multicolumn{4}{c|}{\textbf{OOD Track 1}} & \multicolumn{4}{c|}{\textbf{OOD Track 2}} \\
            \cline{2-13}
             & \textbf{$d_{min}$} & \textbf{$d_{max}$} & \textbf{$\bar{d}$} & \textbf{$\sigma_{d}$} & \textbf{$d_{min}$} & \textbf{$d_{max}$} & \textbf{$\bar{d}$} & \textbf{$\sigma_{d}$} & \textbf{$d_{min}$} & \textbf{$d_{max}$} & \textbf{$\bar{d}$} & \textbf{$\sigma_{d}$} \\
            \hline
            PPO & \textbf{0.0000} & \textbf{0.0283} & \textbf{0.0020} & \textbf{0.0023} & \textbf{0.0003} & \textbf{0.7930} & \textbf{0.3843} & \textbf{0.2252} & \textbf{0.0007} & \textbf{0.9691} & \textbf{0.3384} & \textbf{0.2857} \\ 
            \hline
            SAC & 0.0002 & 1.1637 & 0.4531 & 0.3252 & 0.0030 & 1.8188 & 0.7702 & 0.5459 & $0.0030^{4\dagger}$ & $1.2688^{4\dagger}$ & $0.5447^{4\dagger}$ & $0.3542^{4\dagger}$ \\ 
            \hline
        \end{tabular}%
    }
\end{table*}

\begin{figure*}[!h]
    \centering
    \begin{minipage}{0.45\textwidth}
    \includegraphics[width=\textwidth]{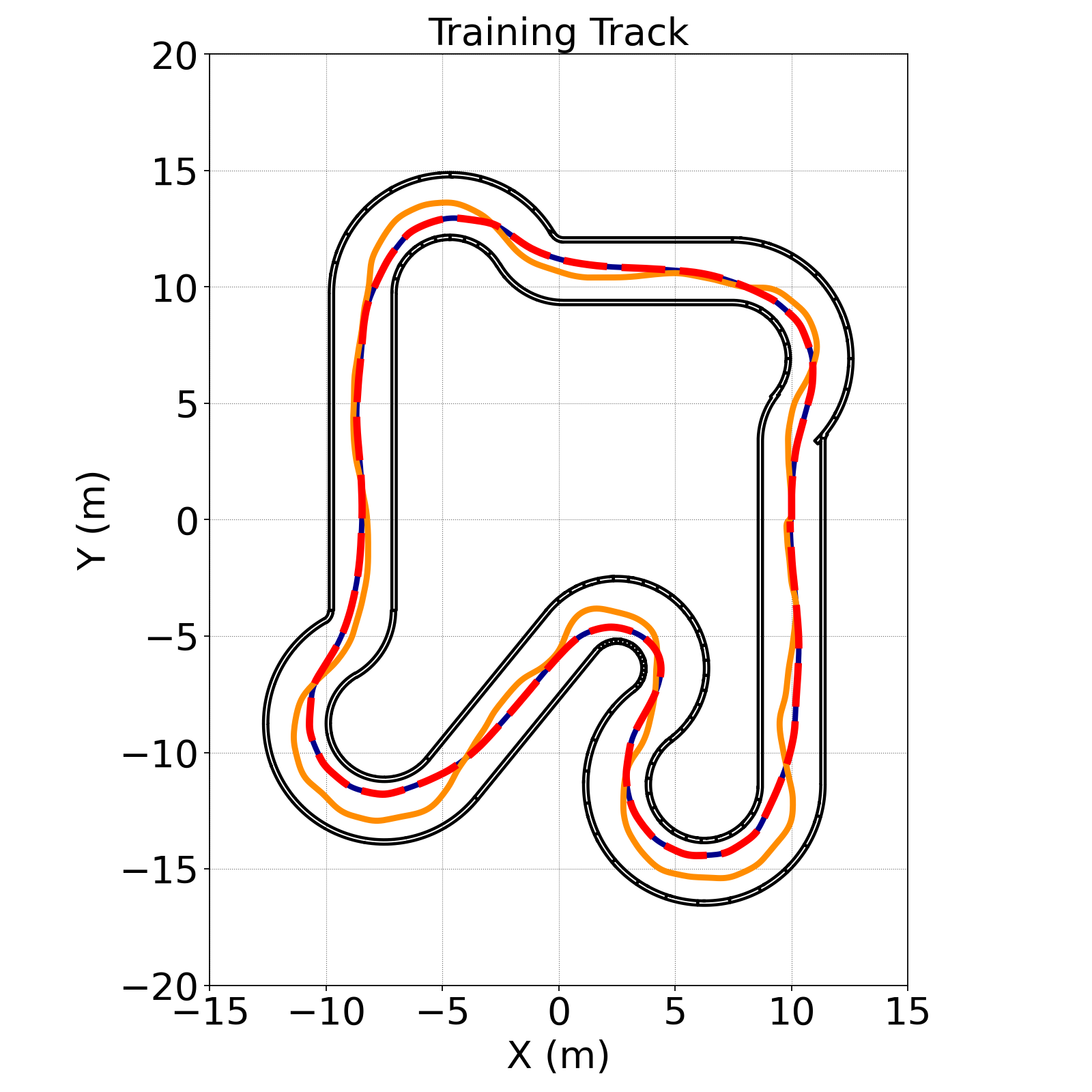}
    \end{minipage}%
    \\
    \begin{minipage}{0.45\textwidth}
    \includegraphics[width=\textwidth]{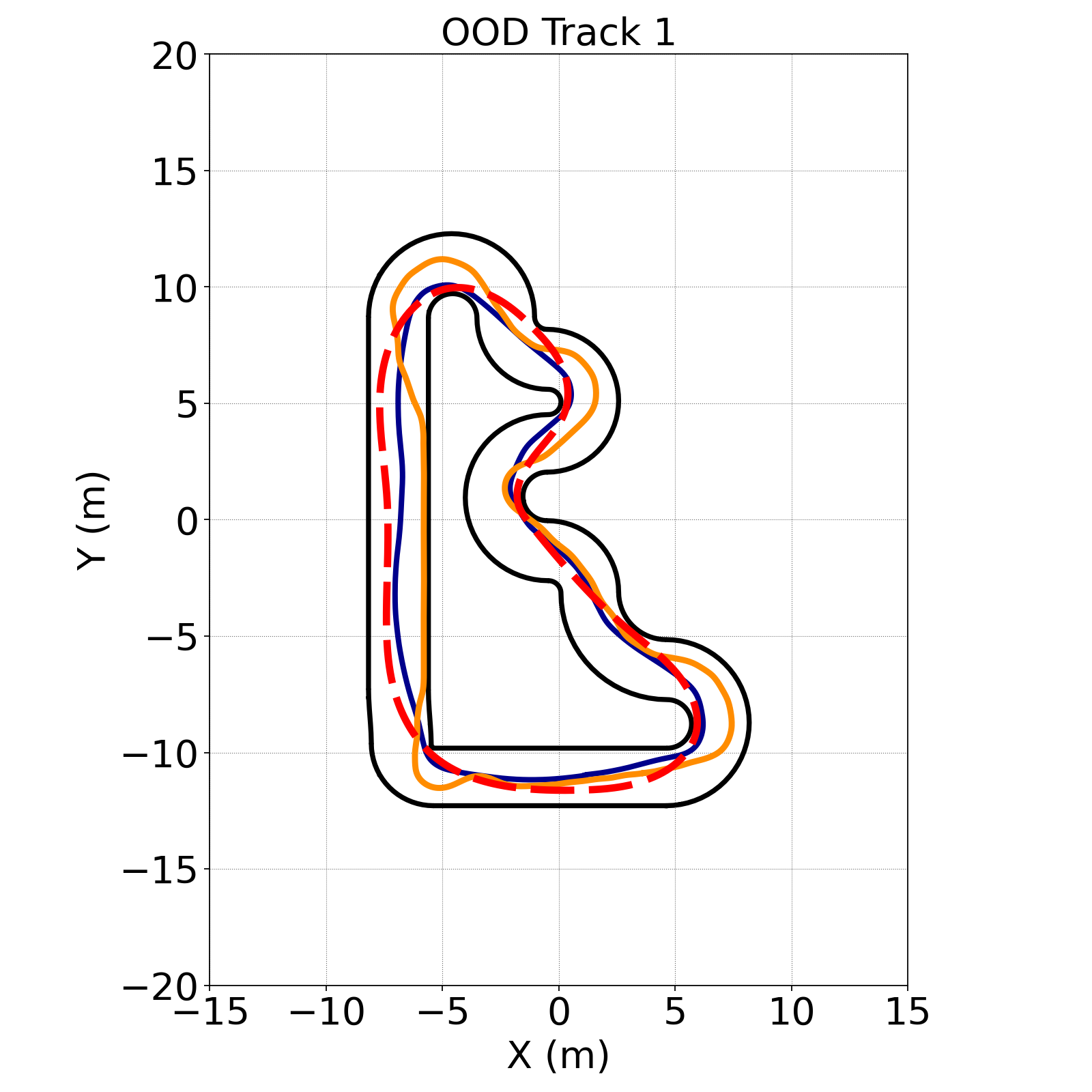}
    \end{minipage}
    \begin{minipage}{0.45\textwidth}
    \includegraphics[width=\textwidth]{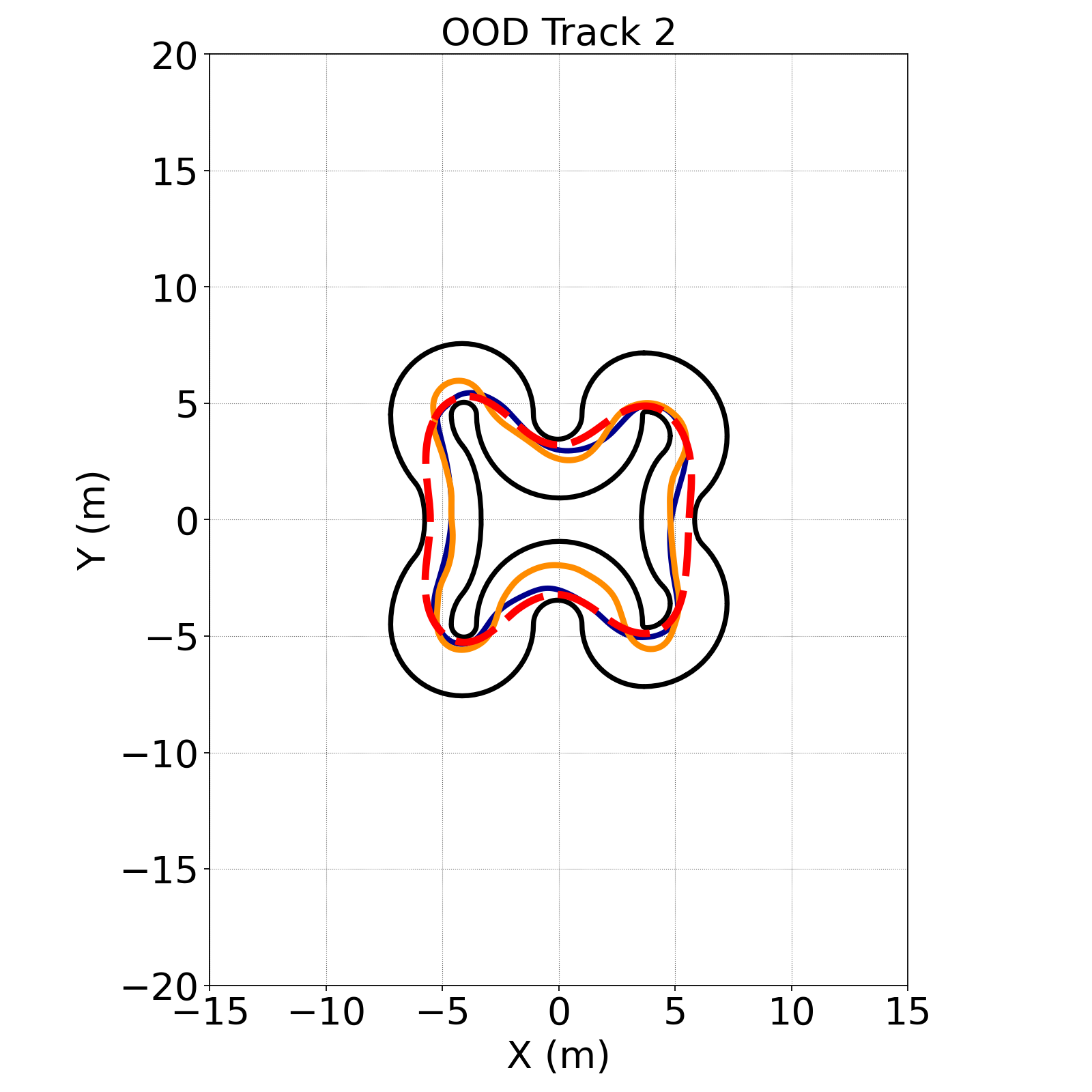}
    \end{minipage}
    \\
    \begin{minipage}{0.15\textwidth}
    \includegraphics[width=\textwidth]{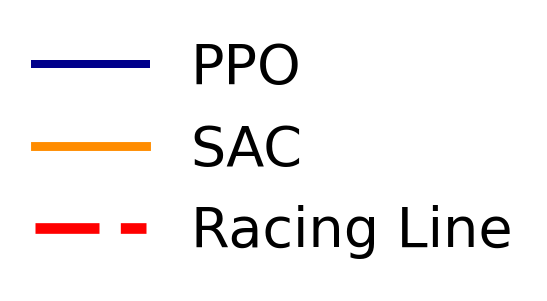}
    \end{minipage}%
    \caption{Trajectories in the simulated tracks.} \label{racetracks_sim_traj}
\end{figure*}

Over a 10-lap evaluation at the curriculum speed which topped out at $1.89 m/s$ with the $c_{T}$ at 20, PPO demonstrated a clear advantage in both kinodynamic speed and trajectory consistency. As summarized in Table \ref{rl_algorithm_lap_times}, PPO posted a 9.00\% faster minimum lap $t_{min}$, a 16.88\% faster maximum lap $t_{max}$, and a 12.16\% faster average time $\bar{t}$ compared to SAC on the training track, while concurrently reducing the standard deviation of lap times $\sigma_{t}$ by 98.01\%.

This performance delta translated effectively to unseen environments without track-specific retraining. In OOD Track 1, PPO outpaced SAC with an 11.48\% faster $t_{min}$, a 15.00\% faster $t_{max}$, and a 13.74\% faster $\bar{t}$, reducing variance by 87.99\%. The generalization drift of off-policy algorithms became evident on OOD Track 2, a more technical configuration than the training track, where SAC failed to reliably generalize and completed only 4 of the 10 laps. PPO successfully finished all 10 laps, achieving a 28.45\% faster $t_{min}$, a 16.33\% faster $t_{max}$ and an 18.61\% faster $\bar{t}$ with 171.51\% increased variance compared to SAC's partial run.

Across all environments, PPO demonstrated significantly tighter adherence to the theoretical baseline with lower minimum $d_{min}$, maximum $d_{max}$ and average $\bar{d}$ lateral deviations that varied less in 10 laps. The trajectories are illustrated in Figure \ref{racetracks_sim_traj} and the lateral deviations are summarized in Table \ref{rl_algorithm_deviation}. On the training track, PPO effectively mirrors the ideal trajectory with a $\bar{d}$ of 0.0020 m, corresponding to a 0.08\% deviation from the mathematical optimum. In contrast, SAC averaged a 0.4531 m error, representing an 18.31\% track width deviation. When evaluated on OOD Track 1, PPO maintained a 15.53\% $\bar{d}$ compared to SAC's 31.12\%. On OOD Track 2, SAC's partial run deviated by 22.01\% on average, whereas PPO deviated by 13.67\%.

This divergence is rooted in the fundamental optimization objectives. High spatial frequency at corner apexes is intrinsically mapped to uncertainty when unconditioned on a geometric reference trajectory, resulting in a conservative safety buffer that repels track boundaries. PPO converges to a deterministic, low-entropy policy to maximize the friction circle for the highest reward, learning to risk near-collisions at the apex, whereas SAC balances entropy maximization, hence the highest reward is not the sole priority, thus collision avoidance to maximize the value horizon converges the objective function.

\begin{table*}[ht]
    \renewcommand{\arraystretch}{1.2}
    \centering
    \caption{Reward Ablation Lap Times ($s$)}
    \label{reward_ablation_lap_times}
    \resizebox{\textwidth}{!}{
        \begin{tabular}{|c|c|c|c|c|c|c|c|c|c|c|c|c|}
            \hline
            \multirow{2}{*}{\textbf{Reward Function}} & \multicolumn{4}{c|}{\textbf{Training Track}} & \multicolumn{4}{c|}{\textbf{OOD Track 1}} & \multicolumn{4}{c|}{\textbf{OOD Track 2}} \\
            \cline{2-13}
             & \textbf{$t_{min}$} & \textbf{$t_{max}$} & \textbf{$\bar{t}$} & \textbf{$\sigma_{t}$} & \textbf{$t_{min}$} & \textbf{$t_{max}$} & \textbf{$\bar{t}$} & \textbf{$\sigma_{t}$} & \textbf{$t_{min}$} & \textbf{$t_{max}$} & \textbf{$\bar{t}$} & \textbf{$\sigma_{t}$} \\
            \hline
            $R$ & \textbf{52.27} & \textbf{52.37} & \textbf{52.33} & \textbf{0.034} & \textbf{33.40} & \textbf{33.63} & \textbf{33.53} & \textbf{0.073} & \textbf{20.03} & \textbf{24.93} & \textbf{23.61} & 1.86 \\
            \hline
            $R_{ab1}$ & 52.70 & 53.27 & 52.90 & 0.16 & 34.30 & 35.03 & 34.67 & 0.238 & 22.67 & 26.3 & 25.11 & \textbf{1.18} \\ 
            \hline
            $R_{ab2}$ & 53.33 & 53.90 & 53.63 & 0.214 & $37.53^{1\dagger}$ & $37.53^{1\dagger}$ & $37.53^{1\dagger}$ & NA & NA & NA & NA & NA\\ 
            \hline
        \end{tabular}%
    }
\end{table*}

\begin{table*}[ht]
    \renewcommand{\arraystretch}{1.2}
    \centering
    \caption{Reward Ablation Deviation from Theoretical Baseline ($m$)}
    \label{reward_ablation_deviation}
    \resizebox{\textwidth}{!}{
        \begin{tabular}{|c|c|c|c|c|c|c|c|c|c|c|c|c|}
            \hline
            \multirow{2}{*}{\textbf{Algorithm}} & \multicolumn{4}{c|}{\textbf{Training Track}} & \multicolumn{4}{c|}{\textbf{OOD Track 1}} & \multicolumn{4}{c|}{\textbf{OOD Track 2}} \\
            \cline{2-13}
             & \textbf{$d_{min}$} & \textbf{$d_{max}$} & \textbf{$\bar{d}$} & \textbf{$\sigma_{d}$} & \textbf{$d_{min}$} & \textbf{$d_{max}$} & \textbf{$\bar{d}$} & \textbf{$\sigma_{d}$} & \textbf{$d_{min}$} & \textbf{$d_{max}$} & \textbf{$\bar{d}$} & \textbf{$\sigma_{d}$} \\
            \hline
            $R$ & \textbf{0.0000} & \textbf{0.0283} & \textbf{0.0020} & \textbf{0.0023} & \textbf{0.0003} & \textbf{0.7930} & \textbf{0.3843} & \textbf{0.2252} & \textbf{0.0007} & \textbf{0.9691} & \textbf{0.3384} & \textbf{0.2857} \\ 
            \hline
            $R_{ab1}$ & \textbf{0.0000} & 0.4960 & 0.1713 & 0.1394 & 0.0006 & 1.4076 & 0.6159 & 0.4185 & 0.0034 & 2.7788 & 1.7650 & 0.7954 \\ 
            \hline
            $R_{ab2}$ & 0.0002 & 0.6317 & 0.2174 & 0.1514 & $0.0003^{1\dagger}$ & $1.9393^{1\dagger}$ & $0.5031^{1\dagger}$ & $0.3422^{1\dagger}$ & NA & NA & NA & NA \\ 
            \hline
        \end{tabular}%
    }
\end{table*}

\begin{table*}[!b]
    \renewcommand{\arraystretch}{1.2}
    \centering
    \caption{Lap Times at the Curriculum and Unconstrained Velocities ($s$)}
    \label{velocity_scaling_lap_times}
    \resizebox{0.9\textwidth}{!}{%
        \begin{tabular}{|c|c|c|c|c|c|c|c|c|c|c|c|c|c|}
            \hline
            \multirow{2}{*}{\textbf{}} & \multicolumn{4}{c|}{\textbf{Training Track}} & \multicolumn{4}{c|}{\textbf{OOD Track 1}} & \multicolumn{4}{c|}{\textbf{OOD Track 2}} \\
            \cline{2-13}
             & \textbf{$t_{min}$} & \textbf{$t_{max}$} & \textbf{$\bar{t}$} & \textbf{$\sigma_{t}$} & \textbf{$t_{min}$} & \textbf{$t_{max}$} & \textbf{$\bar{t}$} & \textbf{$\sigma_{t}$} & \textbf{$t_{min}$} & \textbf{$t_{max}$} & \textbf{$\bar{t}$} & \textbf{$\sigma_{t}$} \\
            \hline
            Curriculum Velocity & 52.27 & 52.37 & 52.33 & \textbf{0.034} & 33.40 & 33.63 & 33.53 & \textbf{0.073} & 20.03 & 24.93 & 23.61 & 1.86 \\ 
            \hline
            Unconstrained Velocity & \textbf{25.20} & \textbf{26.30} & \textbf{25.63} & 0.357 & \textbf{17.63} & \textbf{18.80} & \textbf{18.10} & 0.327 & \textbf{15.17} & \textbf{15.73} & \textbf{15.44} & \textbf{0.169} \\ 
            \hline
        \end{tabular}%
    }
\end{table*}

\textbf{Reward Ablation.} To validate the necessity of the proposed reward formulation $R$, ablation studies were conducted against an unsquared throttle $R_{ab1}$ and a collision penalty $R_{ab2}$ at the curriculum speed, evaluated over 10 laps. $R$ parameterized the best policy within 20,000,000 time steps, while the ablated rewards plateaued by 10,000,000. As detailed in Tables \ref{reward_ablation_lap_times} and \ref{reward_ablation_deviation}, $R$ held an advantage over the ablations in lap times and deviations from the racing line across all tracks.

In the training track $R$ lapped 1.07\% faster on average than $R_{ab1}$ and tightened lap consistency by 78.93\%. Introducing a collision penalty with $R_{ab2}$ injected conservatism, decreasing performance across all time metrics by up to 2.84\% and increased variance by 84.11\%. Furthermore, $R_{ab2}$ failed to generalize, managing only a single lap on OOD Track 1 and failed on the technical OOD Track 2. On the other hand, $R$ lapped 3.29\% faster on average than $R_{ab1}$ in OOD Track 1 with 69.33\% lower variance, and posted $t_{min}$ 11.01\% faster than the sole lap by $R_{ab2}$. In OOD Track 2, $R$ had 56.83\% more variance than $R_{ab1}$ due to an 11.62\% faster $t_{min}$ that was substantially faster than the $\bar{t}$ and $t_{max}$ reductions of 5.97\% and 5.20\%.

Analyzing the trajectory precision further corroborates the importance of promoting $T$ nonlinearly by squaring it, and eliminating an explicit collision penalty. While $R$ mirrored the minimum curvature racing line with a 0.08\% average deviation in the training track, $R_{ab1}$ and $R_{ab2}$ deviated 6.92\% and 8.78\% on average. In OOD Tracks 1 and 2, $R_{ab1}$ deviated by 24.89\% and 71.32\% on average, substantially deviating in the latter that was most out of the training distribution, whereas $R$ deviated by 15.53\% and 13.67\%. The sole lap with $R_{ab2}$ in OOD Track 1 deviated by 20.33\%, 4.80\% more than $R$ on average. $R$ completed the laps with the least lateral deviation variance in all tracks.

\textbf{Velocity Scaling. } PPO trained with $R$ at the curriculum velocity with $C_{T} = 20$ was deployed zero-shot at the maximum hardware velocity with $C_{T} = 120$ to evaluate the policy's capacity to extrapolate encoded dynamics for kinodynamic stability at unconstrained physical limits.

\begin{table*}[ht]
    \renewcommand{\arraystretch}{1.2}
    \centering
    \caption{Velocities at the Curriculum and Unconstrained Velocities ($m/s$)}
    \label{velocity_scaling_velocities}
    \resizebox{0.8\textwidth}{!}{%
        \begin{tabular}{|c|c|c|c|c|c|c|c|c|c|}
            \hline
            \multirow{2}{*}{\textbf{}} & \multicolumn{3}{c|}{\textbf{Training Track}} & \multicolumn{3}{c|}{\textbf{OOD Track 1}} & \multicolumn{3}{c|}{\textbf{OOD Track 2}} \\
            \cline{2-10}
             & \textbf{$v_{min}$} & \textbf{$v_{max}$} & \textbf{$\bar{v}$} & \textbf{$v_{min}$} & \textbf{$v_{max}$} & \textbf{$\bar{v}$} & \textbf{$v_{min}$} & \textbf{$v_{max}$} & \textbf{$\bar{v}$}\\
            \hline
            Curriculum Velocity & 1.571 & 1.888 & 1.806 & \textbf{1.583} & 1.880 & 1.789 & \textbf{1.496} & 1.889 & 1.757 \\ 
            \hline
            Unconstrained Velocity & \textbf{1.624} & \textbf{4.832} & \textbf{3.755} & 1.3858 & \textbf{4.8650} & \textbf{3.4594} & 1.2183 & \textbf{4.5675} & \textbf{3.0137} \\ 
            \hline
            
        \end{tabular}%
    }
\end{table*}

\begin{figure*}[!h]
    \centering
    \begin{minipage}{0.45\textwidth}
    \includegraphics[width=\textwidth]{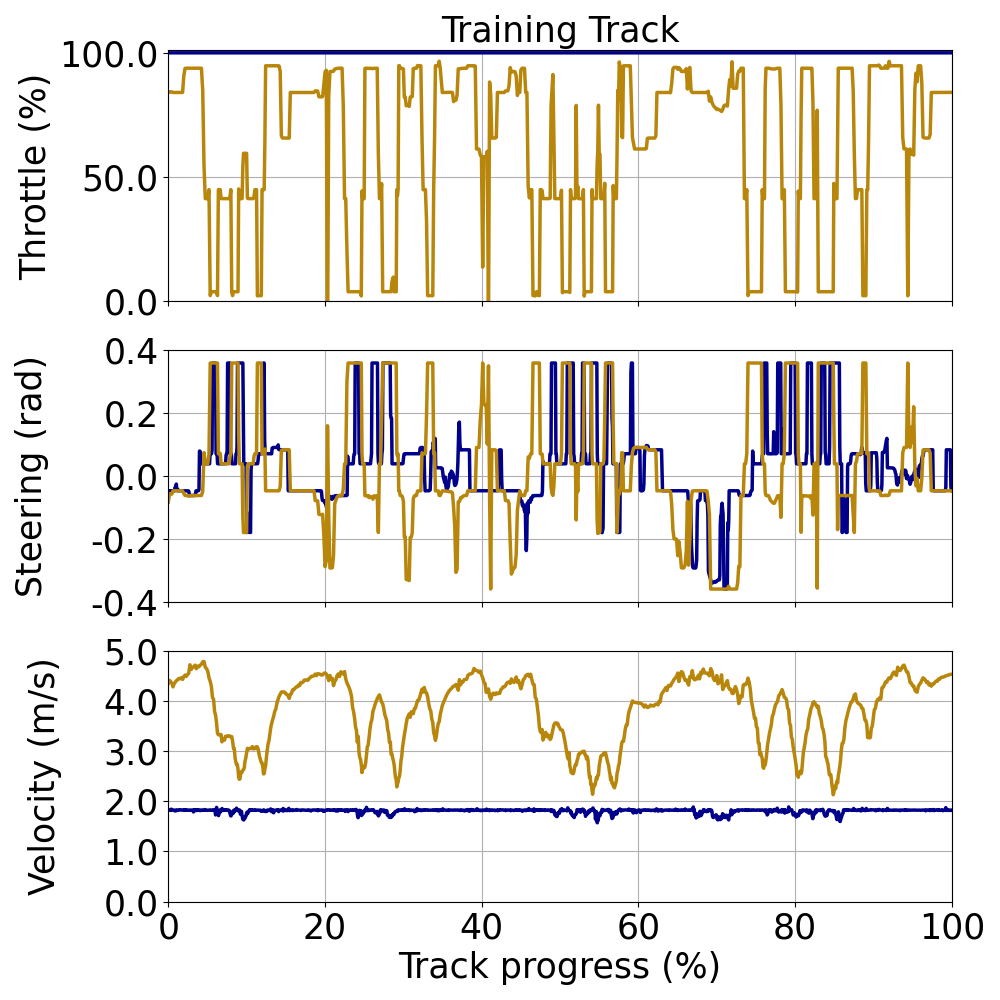}
    \label{racetrack1_telem}
    \end{minipage}%
    \\
    \begin{minipage}{0.45\textwidth}
    \includegraphics[width=\textwidth]{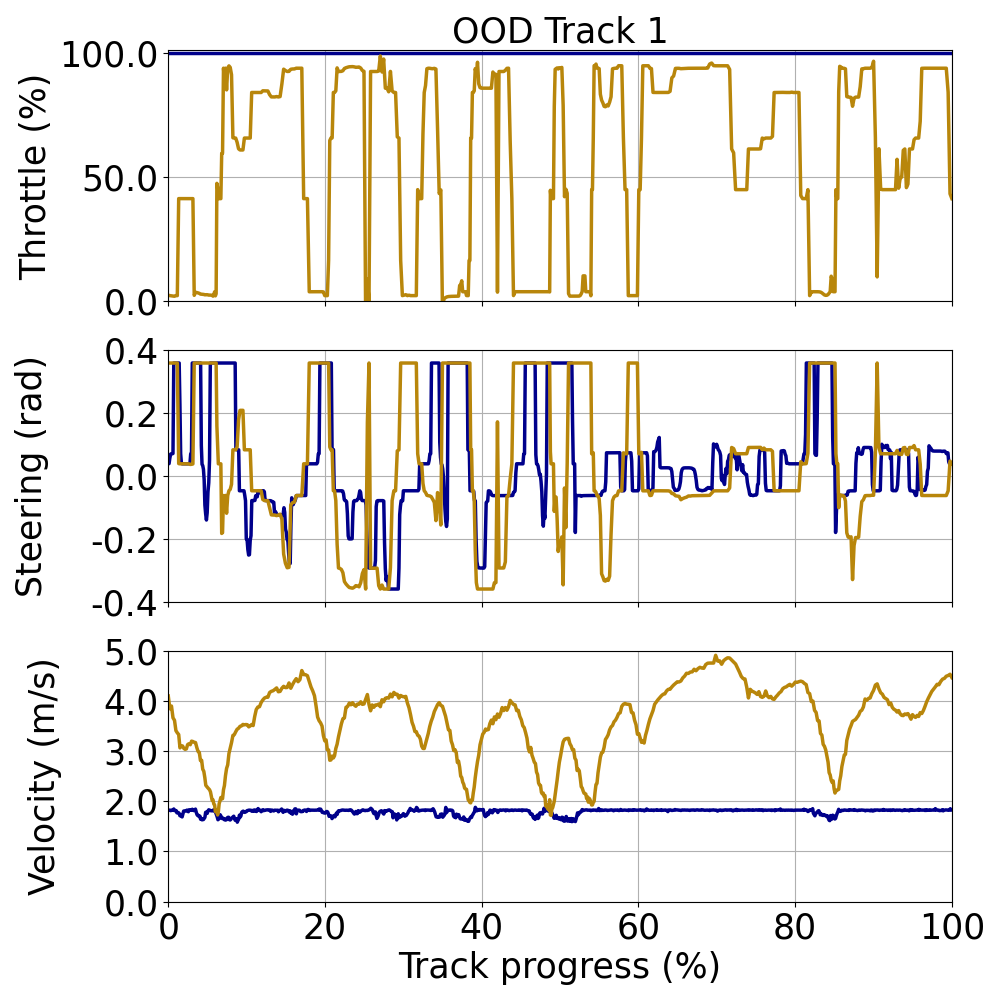}
    \label{racetrack2_telem}
    \end{minipage}
    \begin{minipage}{0.45\textwidth}
    \includegraphics[width=\textwidth]{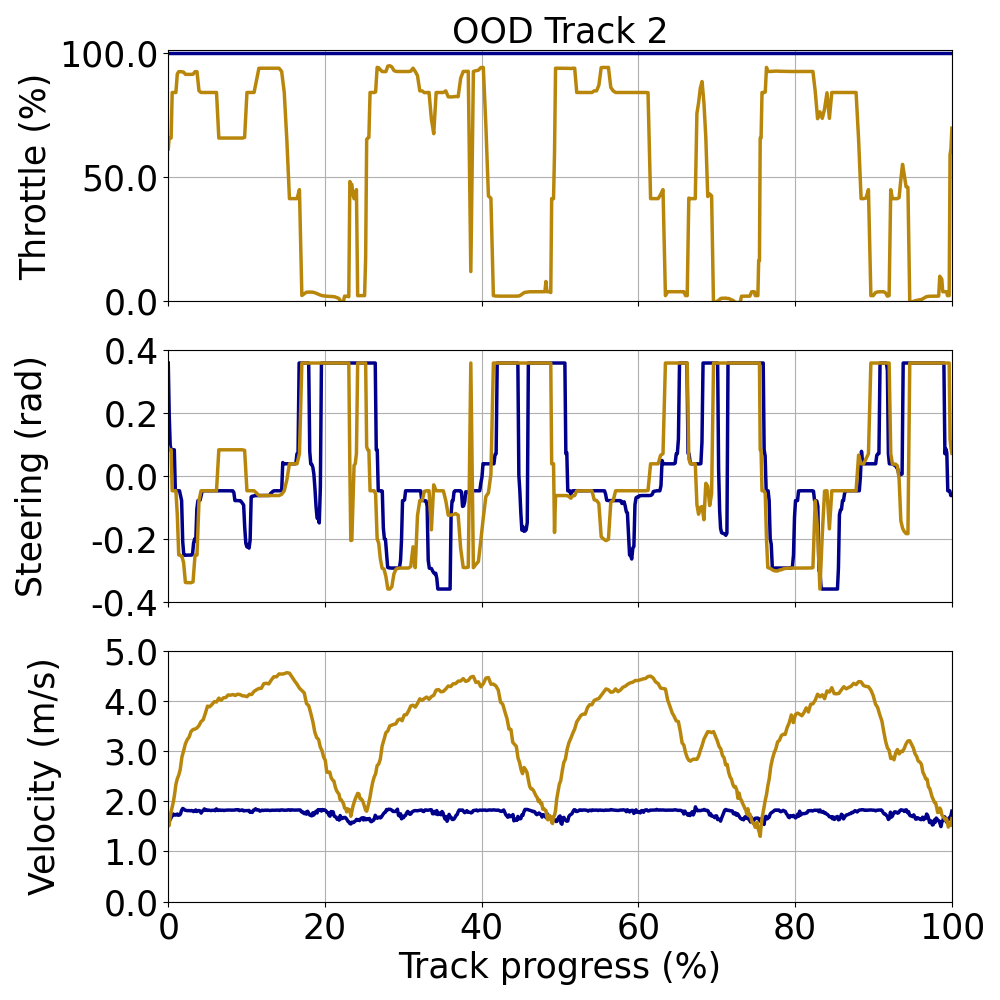}
    \label{racetrack3_telem}
    \end{minipage}
    \\
    \begin{minipage}{0.5\textwidth}
    \includegraphics[width=\textwidth]{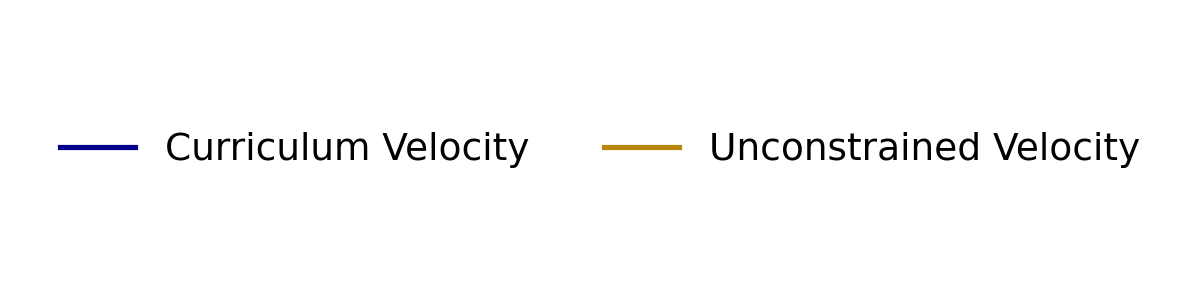}
    \end{minipage}%
    \caption{Telemetry at the curriculum and unconstrained velocities.} \label{racetracks_sim_telemetry}
\end{figure*}

This resulted in substantially faster lap completion times and velocities over 10 laps, summarized in Tables \ref{velocity_scaling_lap_times} and \ref{velocity_scaling_velocities}. On the training track, the unconstrained velocity yielded a 51.79\% faster $t_{min}$, a 49.78\% faster $t_{max}$, and a 51.02\% faster $\bar{t}$. Concurrently, the standard deviation of lap times increased by 950.00\%, managing variance within four tenths of a second. The average velocity $\bar{v}$ increased by 107.92\%, the maximum velocity $v_{max}$ by 155.93\%, and the minimum velocity $v_{min}$ by 3.37\%. 

This performance delta translated effectively to unseen environments without track-specific retraining. In OOD Track 1, the policy utilized the uncapped velocity to outpace the curriculum restriction with a 47.21\% faster $t_{min}$, a 44.10\% faster $t_{max}$, and a 46.01\% faster $\bar{t}$ with an increase in variance of 348.40\%. In the technical OOD Track 2, it produced more consistent laps than at the curriculum speed, with a variance reduction of 90.89\% while posting a 24.29\% faster $t_{min}$, a 36.90\% faster $t_{max}$ and a 34.60\% faster $\bar{t}$. In OOD Track 1, $v_{max}$ increased by 158.78\% and $\bar{v}$ by 93.37\%. In contrast to the training track, the $v_{min}$ decreased by 12.46\%. This divergence in $v_{min}$ is mirrored in OOD Track 2, where $v_{max}$ increased by 141.79\% and $\bar{v}$ by 71.53\%, but $v_{min}$ dropped by 18.56\%.

The reduction in $v_{min}$ at the unconstrained speed reflects the necessity to manage higher entry momentum. At the curriculum speed, this is capped, enabling a higher speed through corners without exceeding the limits of the tire friction circle. When unconstrained, these geometric features are approached at higher velocities, thus to navigate the apex, the policy drops the throttle to trail-brake. This maintains the vehicle state within permissible dynamic boundaries, resulting in a lower instantaneous minimum velocity mid-corner but with a higher momentum profile than at the curriculum speed.

The controls and velocity of the policy at the curriculum and unconstrained speeds in each track are shown in Figure \ref{racetracks_sim_telemetry}. In all tracks, the policy navigates with 100\% throttle at the curriculum speed with a stable velocity profile, and modulates substantially at the unconstrained speed, dropping from 100\% to 0\% to maximize momentum with fluctuating velocities. The steering controls are more similar, adhering to the optimum path, with the inputs at a greater amplitude on average at the unconstrained speed to handle the increased momentum.  

\subsection{Zero-Shot Physical Transfer}

\begin{figure}
\centering
\begin{minipage}{0.252\textwidth}
\includegraphics[width=0.8\textwidth]{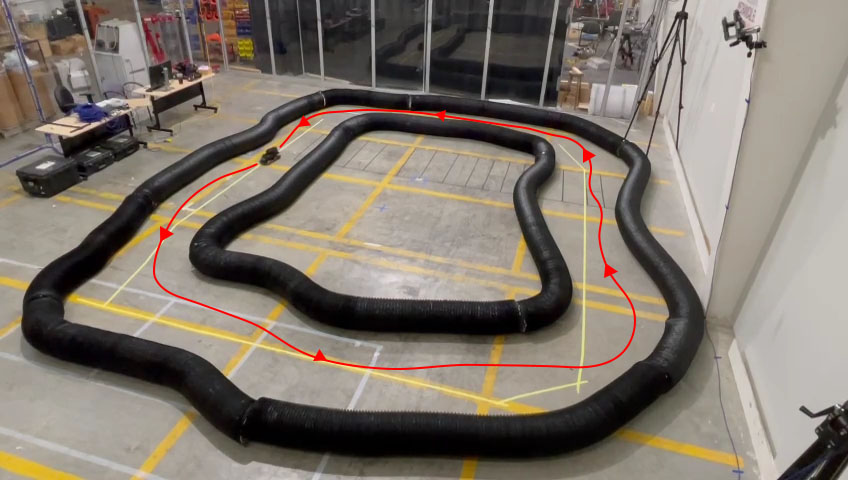}
\subcaption{} \label{racetrack1_traj}
\end{minipage}%
\begin{minipage}{0.252\textwidth}
\includegraphics[width=0.8\textwidth]{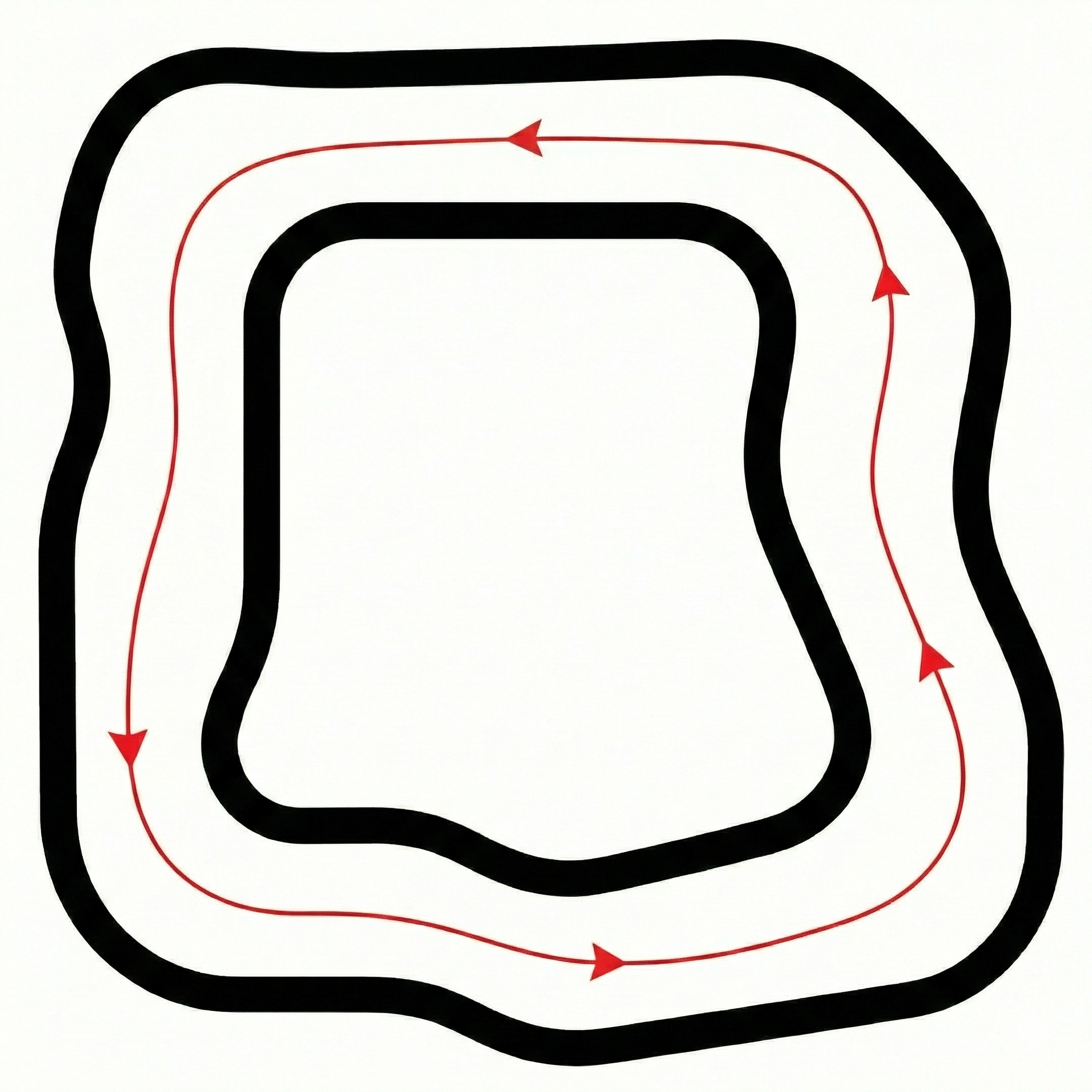}
\end{minipage}
\\
\begin{minipage}{0.252\textwidth}
\includegraphics[width=0.8\textwidth]{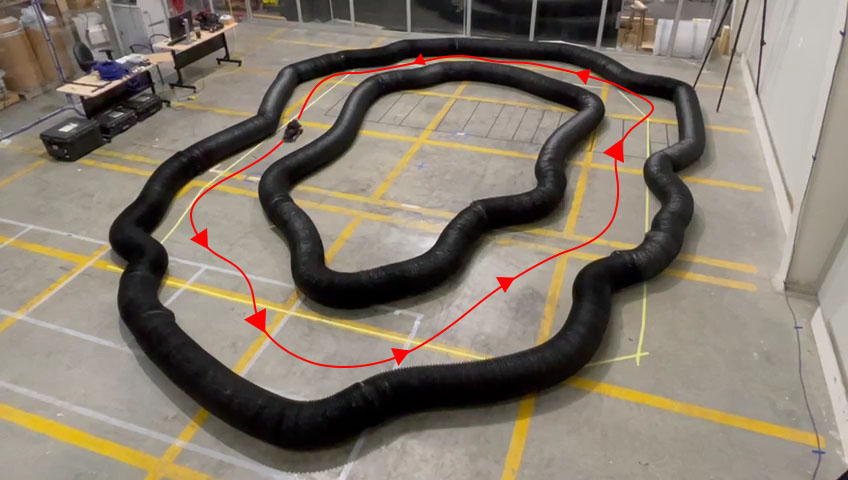}
\subcaption{} \label{racetrack3_traj}
\end{minipage}%
\begin{minipage}{0.252\textwidth}
\includegraphics[width=0.8\textwidth]{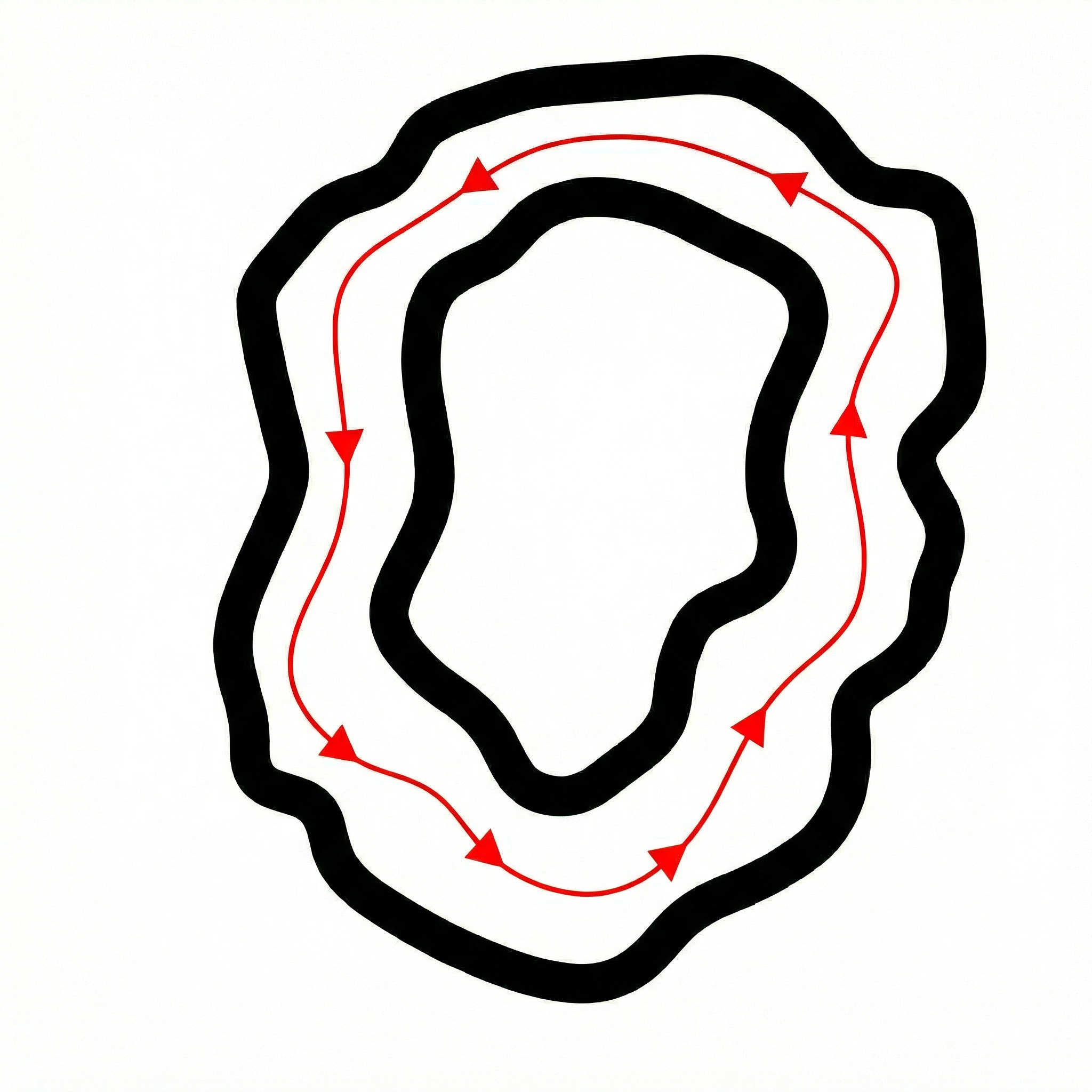}
\end{minipage}
\\
\begin{minipage}{0.252\textwidth}
\includegraphics[width=0.8\textwidth]{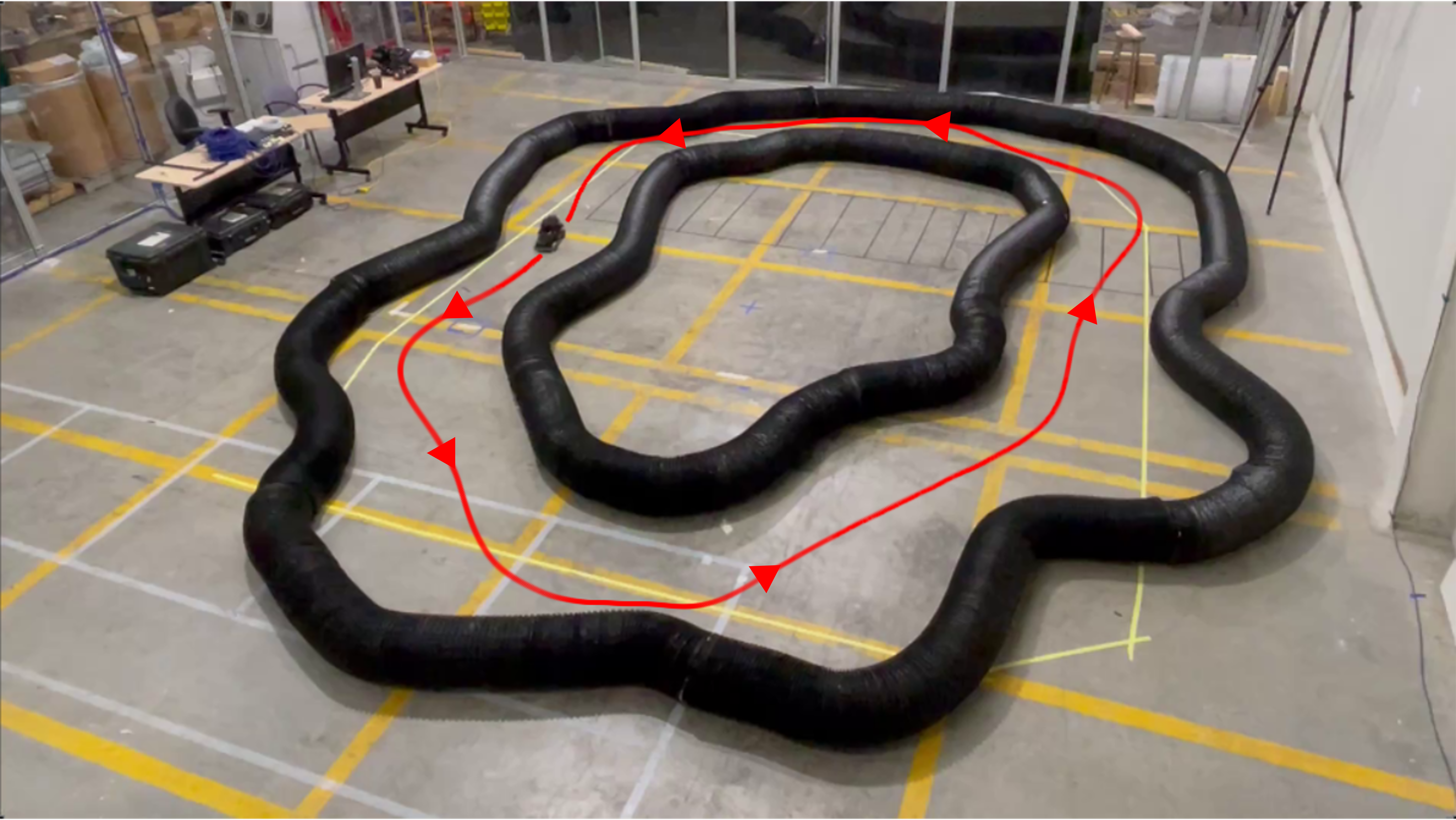}
\subcaption{} \label{racetrack5_traj}
\end{minipage}%
\begin{minipage}{0.252\textwidth}
\includegraphics[width=0.8\textwidth]{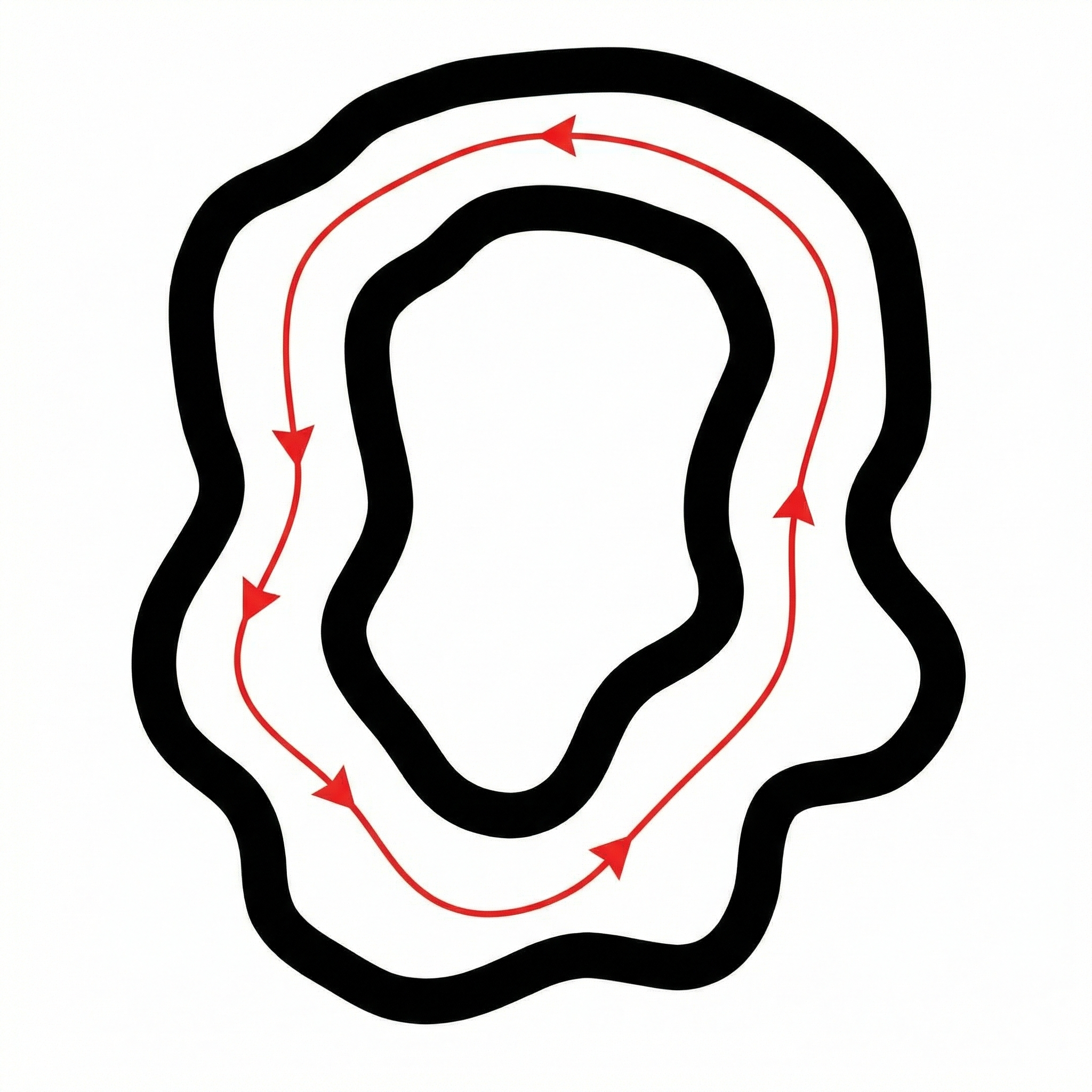}
\end{minipage}
\caption{Zero-shot trajectories in OOD physical track configurations. A video can be viewed at \href{https://www.youtube.com/watch?v=DVxlOARi4aY}{https://www.youtube.com/watch?v=DVxlOARi4aY}.} \label{racetracks_traj}
\end{figure}

Zero-shot sim-to-real transfer capabilities were evaluated on a 1/10th proportionally scaled vehicle \cite{sivashangaran2023xtenth} across unseen tracks, as illustrated in Figure \ref{racetracks_traj}. A Nonlinear Predictive Geometric PID, that we used in the 2023 IEEE Intelligent Vehicles Symposium F1TENTH Competition, and a human demonstration that represents a best case BC performance were utilized as benchmarks.

The competition controller is a baseline reference that was the sole method to complete 10 collision-free laps during qualifying and outperforms the standard Follow-the-Gap (FTG) algorithm. While FTG is reactive when a spatial gap physically appears in the depth scan, often resulting in late apexing, this map independent autonomous baseline utilizes geometric projected look-ahead and wall angle parameters to predict and plan in advance, for higher entry momentum. Moreover, it implements bilateral constraint switching, dynamically anchoring its tracking reference to the nearest track boundary rather than averaging free space, and solves slaloming, common in reactive planners, utilizing gain scheduling with hysteresis. This nonlinear control law dampens oscillations on straights with low gains while quadrupling stiffness during sustained cornering, eliminating the decision-flickering characteristic of FTG. The average lap times with each method over 10 laps are summarized in Table \ref{laptimes}.

\begin{table}[ht]
    \renewcommand{\arraystretch}{1.2}
    \centering
    \caption{Physical Track Lap Times (s)}
    \label{laptimes}
        \resizebox{\columnwidth}{!}{
        \begin{tabular}[t]{|c|c|c|c|c|c|}
            \hline
            \textbf{Method} & \textbf{Track(a)} & \textbf{Track(b)} & \textbf{Track(c)}\\
            \hline
            Human Demonstration & 10.84 & 10.54 & 10.30\\
            \hline
            Nonlinear Predictive Geometric PID  & 12.84 & 12.56 & 12.38\\
            \hline
            \textbf{Sim-to-Real DRL} & \textbf{9.56} & \textbf{9.16} & \textbf{9.10} \\
            \hline
        \end{tabular}
        }
\end{table}

BC represents the SOTA for mapless end-to-end ML navigation, with a theoretical maximum performance bounded by the human demonstrator, which the DRL policy outperforms by 12\%. The gap is more than double that to the classical geometric controller, with a 26\% faster lap time. The faster times and OOD generalization are attributed to the rate of corrections during every segment of the track, some of which occur mid-corner, that are arduous with human reaction times and require dense track-specific tuning with classical controllers. While trajectory-tracking DRL methods achieve zero-shot sim-to-real transfer, these require prebuilt maps and precomputed reference trajectories, whereas our method adapts to each OOD track, requiring only instantaneous spatial measurements.

\begin{figure}
    \centering
    \includegraphics[width = 0.9\columnwidth]
    {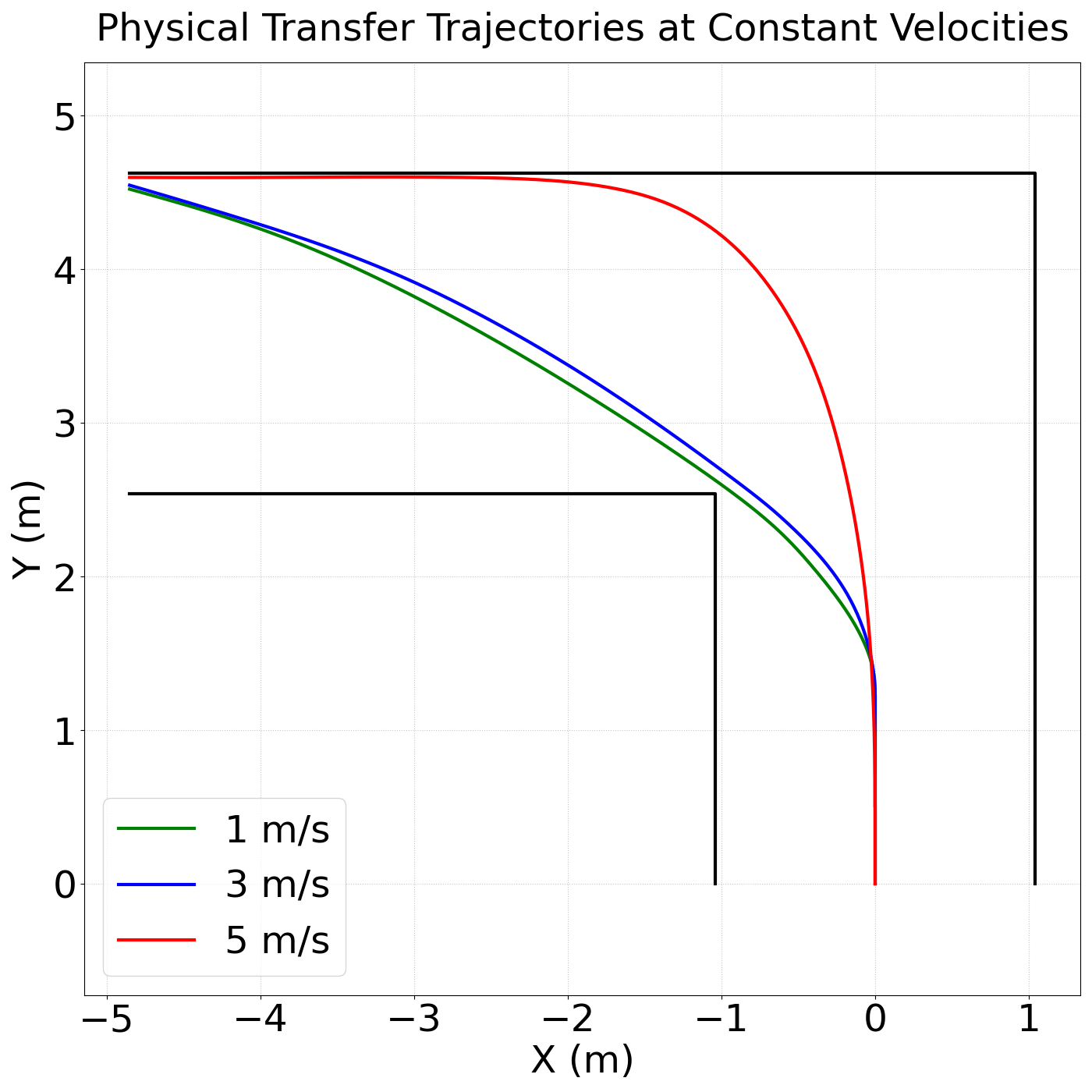}
    \caption{Trajectories on hardware at different constant velocities.} \label{plot_trajectories_velocities}
\end{figure}

The top speed in these tracks is 2 $m/s$, with throttle control to maximize momentum without collision. To validate sim-to-real transfer without slaloming at high-speeds, we tested navigation around a $90^\circ$ corner at constant 1, 3 and 5 $m/s$. The trajectories are illustrated in Figure \ref{plot_trajectories_velocities}. 

At all velocities, the scaled car navigates the corner, that has an angle not in the training track, smoothly, without actuation instability. The trajectories at 1 and 3 $m/s$ are near identical, minimizing distance traveled by turning into the corner apex and utilizing the track width on the exit to carry momentum. At 3 $m/s$, the trajectory is wider than at 1 $m/s$ due to the greater tire friction. The lateral momentum exceeds the tire friction limit at 5 $m/s$ resulting in understeer that misses the apex, and carries maximum momentum toward the outside of the track.

The velocity and time to navigate the corner show a nonlinear relationship with diminishing returns. Increasing the vehicle's velocity from 1 to 3 $m/s$ resulted in a 63.77\% reduction in lap time, from 6.9 $s$ to 2.5 $s$. Conversely, further acceleration to 5 $m/s$ yielded an improvement of 0.5 $s$, a 20.00\% decrease. At lower velocities, the proportion of time spent traversing the straights is high, enabling longitudinal speed increases to significantly impact overall performance. As peak velocity rises, the duration in the straights becomes negligible compared to deceleration, apex navigation, and exit acceleration due to the physical constraints of the track geometry and the dynamic limits of the vehicle. Consequently, the overall track performance becomes fundamentally limited by the maximum lateral grip of the tires. Although past the limits of optimal grip at 5 $m/s$, the RL policy completes the turn and resumes stable straight-line navigation without oscillating.

\subsection{Computational Complexity and Inference Efficiency}

To quantify the compute efficiency, we evaluate the Multiply-Accumulate operations (MACs) per inference step for the proposed framework against SOTA model-based and BC architectures. A MAC operation consists of computing the product of two numbers and adding that result to an accumulator, serving as a fundamental metric for evaluating neural network computational complexity and hardware latency. A summary of the computational requirements is tabulated in Table \ref{mac_comparison}.

\begin{table}[ht]
    \renewcommand{\arraystretch}{1.2}
    \centering
    \caption{Comparison of ML Model Multiply-Accumulate Operations}
    \label{mac_comparison}
        \resizebox{\columnwidth}{!}{
        \begin{tabular}[t]{|c|c|c|}
            \hline
            \textbf{Method} & \textbf{Architecture} & \textbf{MACs}\\
            \hline
            Model-Based DRL \cite{brunnbauer2022latent} & RNN \& MLP & 1,615,200\\
            \hline
            BC \cite{zarrar2024tinylidarnet} &  1D CNN & 240,752 - 1,546,960\\
            \hline
            \textbf{Model-Free DRL} & \textbf{MLP} & \textbf{15,104}\\
            \hline
        \end{tabular}
        }
\end{table}

Model-based DRL agents have a larger computational footprint than model-free, due to latent state-space processing \cite{brunnbauer2022latent, Hafner2020Dream}. In Dreamer, the active inference policy is a recurrent pipeline comprising a 400-node representation model, a 400-unit Gated Recurrent Unit (GRU) transition model, and a 4-layer actor network with 400 neurons per layer. Executing this continuous architecture necessitates approximately 1.62 million MACs per step, during which the representation model compresses a high-dimensional 2D LiDAR scan into a 30-dimensional stochastic latent state, following which the GRU updates its 400-dimensional hidden state using a 32-dimensional concatenated vector, yielding a 430-dimensional combined latent state that drives the actor network.

BC architectures, such as TinyLidarNet \cite{zarrar2024tinylidarnet}, utilize 1D CNNs to extract spatial features, where the computational cost scales directly with the input resolution and filter depth. The full model processes the complete range of depth sensor measurements, resulting in 1,546,960 MACs. By reducing the network depth and filter dimensions, the small variant trims this requirement to 240,752 MACs, compromising track performance for feasibility in a range of embedded systems.

Model-free DRL architectures that downsample depth measurements to 20 rays require significantly fewer MACs and outperform model-based and BC in simulation, but consistently exhibit high-frequency actuation instability during sim-to-real transfer. These comprise CNNs \cite{bosello2022train} which require 51,520 MACs and MLPs \cite{evans2023comparing} that require 14,300. By solving sim-to-real slaloming, we encode the nonlinear vehicle dynamics with a spatial resolution of 170 using large-scale simulation accelerated training, in a MLP with 2 hidden layers of 64 nodes each, and an output layer of 2 continuous actions, that requires 15,104 MACs. This formulation maintains a computational overhead comparable to the downsampled and unstable model-free baselines, and requires less than 1\% of the footprint of SOTA model-based and BC architectures. 

\subsection{Dynamic Multi-Agent Overtaking}

\begin{figure*}
    \centering
    \begin{minipage}{0.45\textwidth}
    \includegraphics[width=\textwidth]{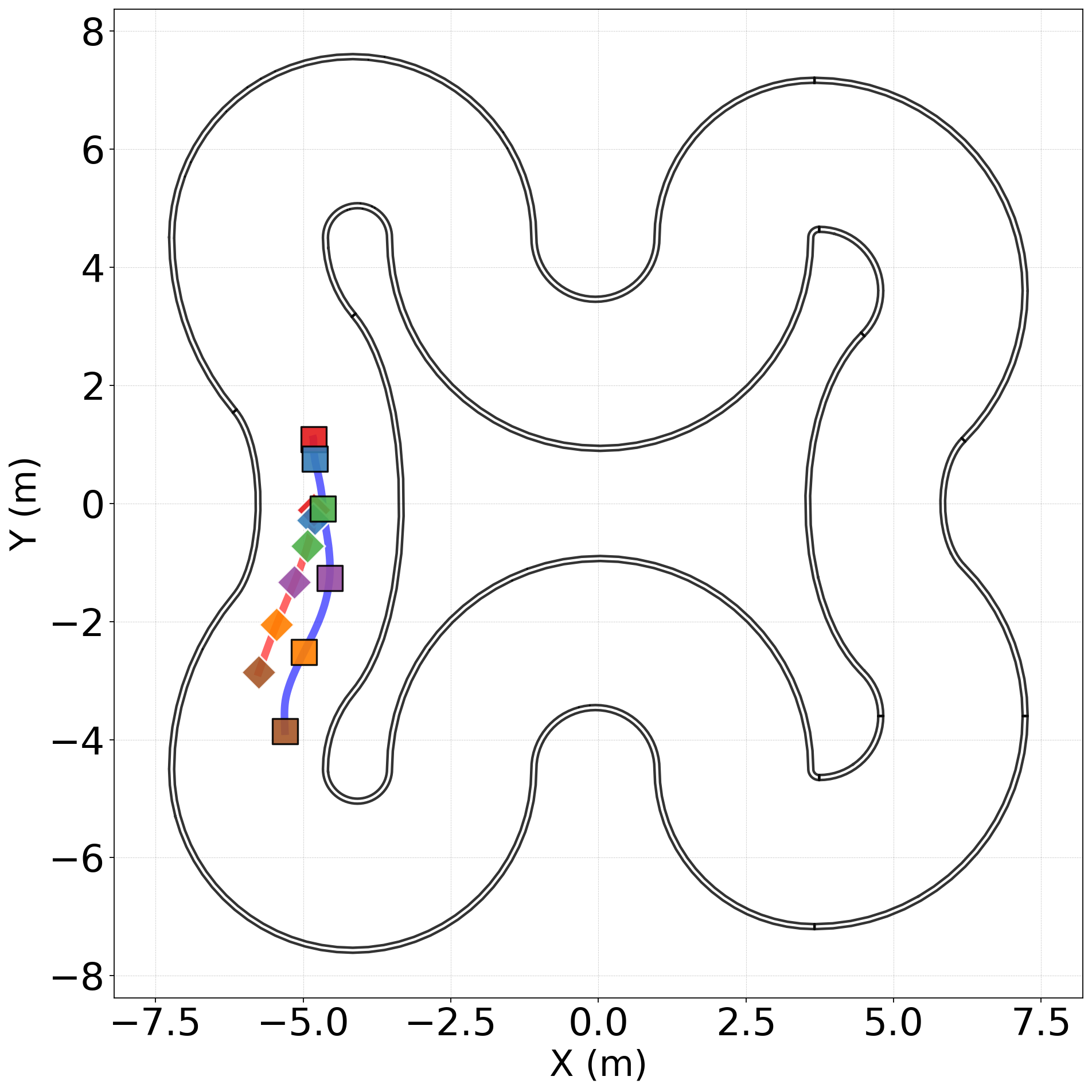}
    \end{minipage}%
    \\
    \begin{minipage}{0.45\textwidth}
    \includegraphics[width=\textwidth]{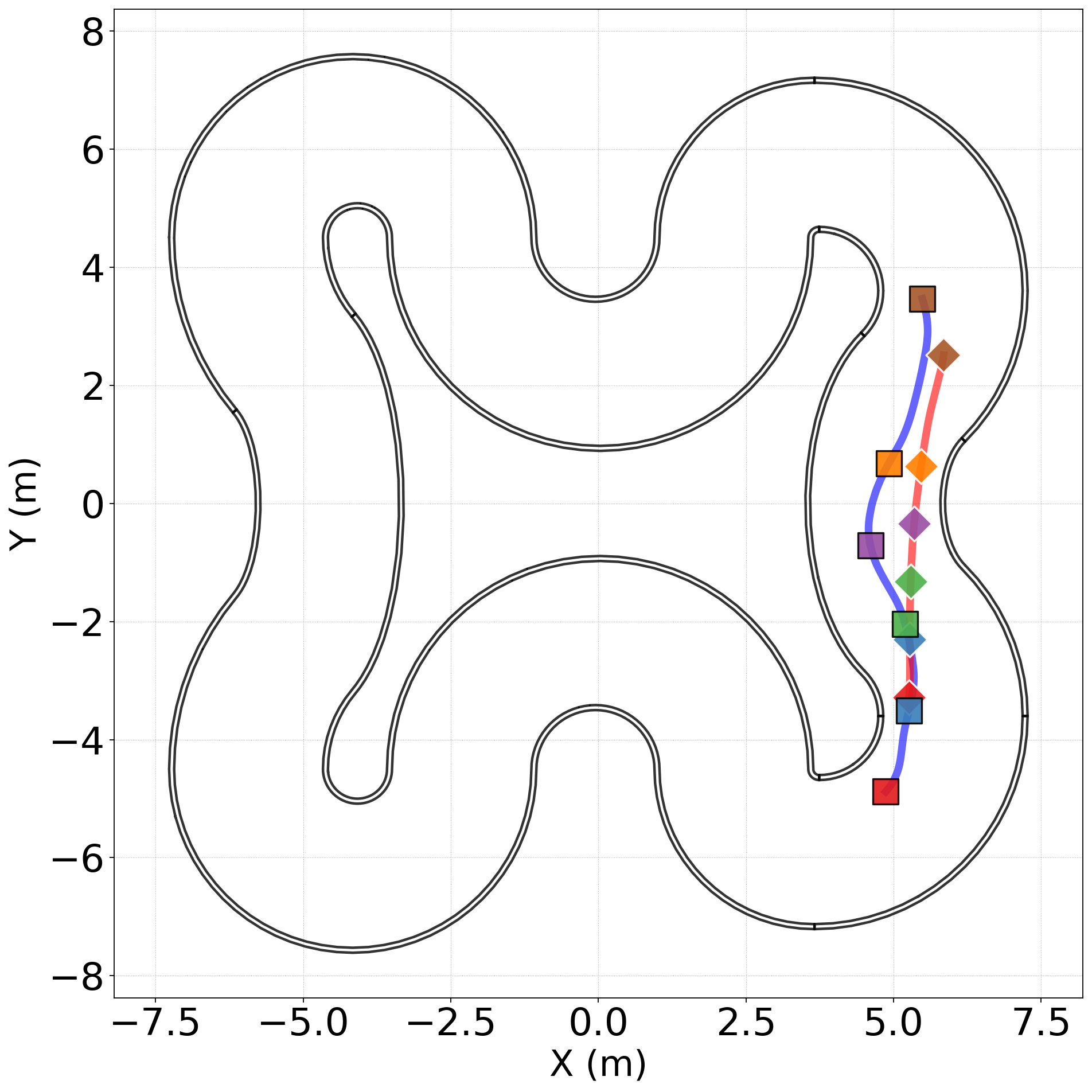}
    \end{minipage}
    \begin{minipage}{0.45\textwidth}
    \includegraphics[width=\textwidth]{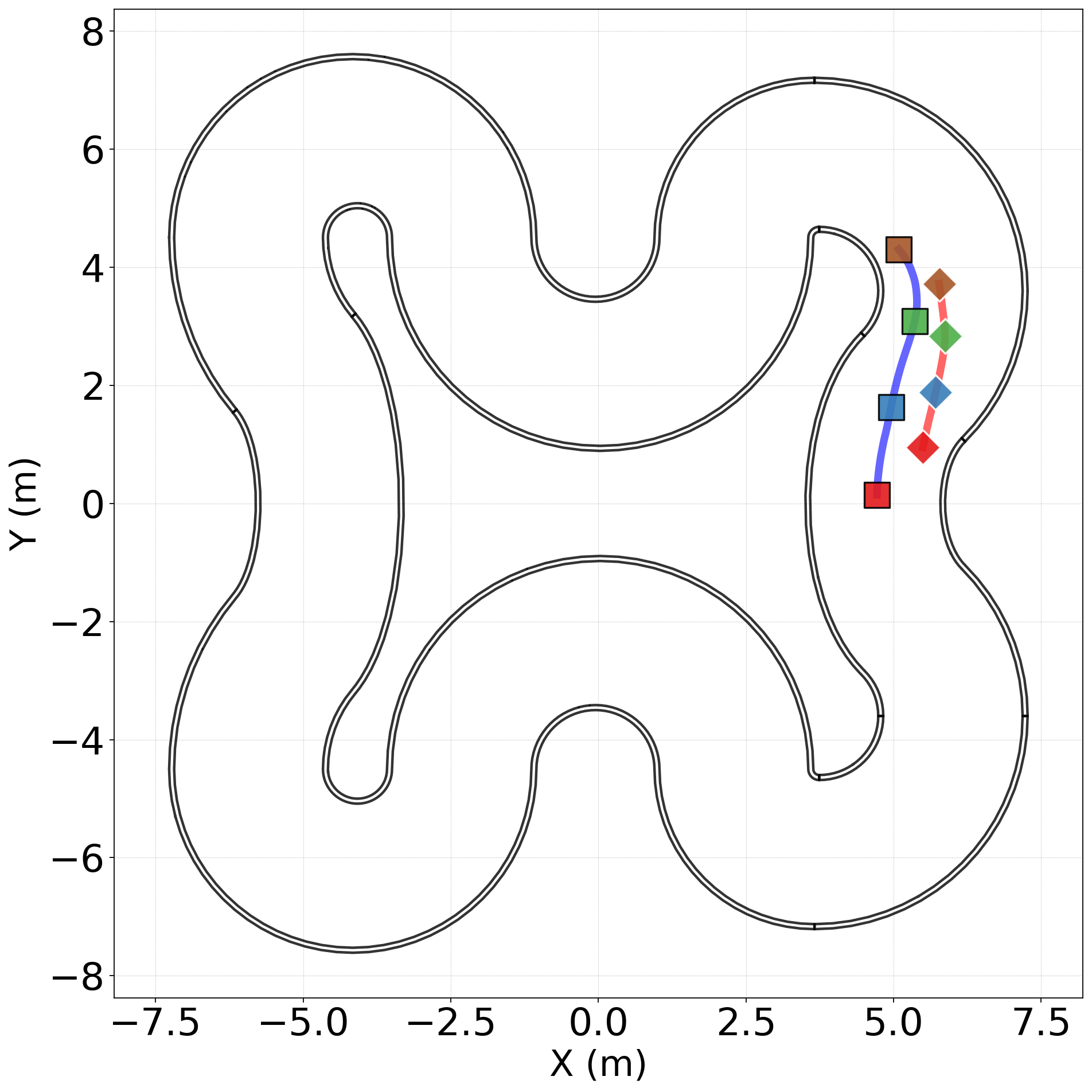}
    \end{minipage}
    \\
    \begin{minipage}{0.4\textwidth}
    \includegraphics[width=\textwidth]{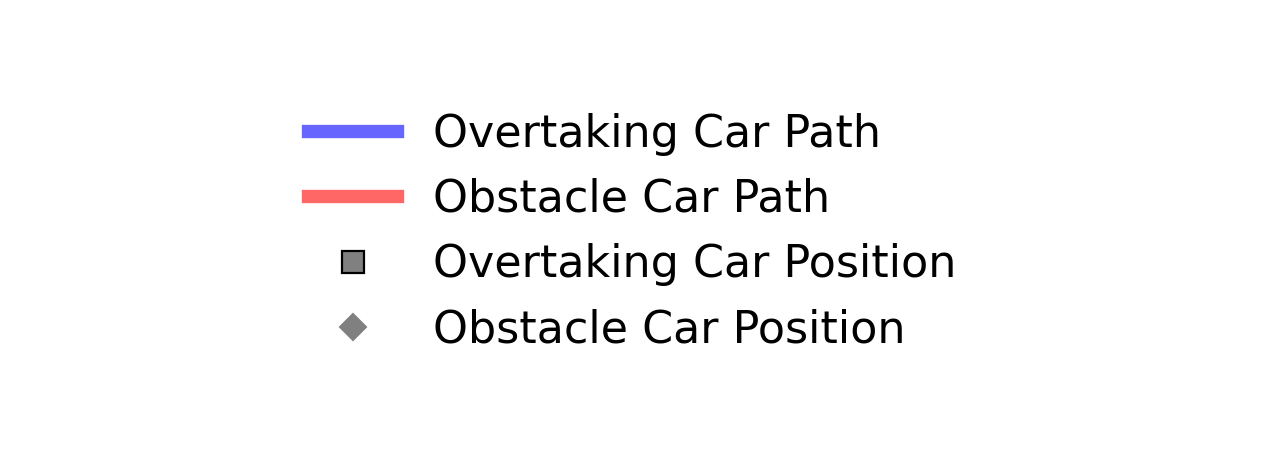}
    \end{minipage}%
    \caption{Examples of overtaking trajectories. The positions of the overtaking and obstacle cars are causally color coded.} \label{plot_ot_traj}
\end{figure*}

The adaptability of the physics-informed dynamics encoding facilitates overtaking without specialized multi-agent algorithmic formulations and precomputed reference trajectories. The policy was deployed in a dynamic multi-agent environment configured within OOD Track 2 which comprises fifteen obstacle vehicles navigating the track utilizing the pretrained DRL policy. To facilitate overtaking, the ego vehicle was assigned a $C_{T}$ doubled in magnitude, providing a kinodynamic acceleration advantage.

\begin{figure}
\centering
\begin{minipage}{0.252\textwidth}
\includegraphics[width=0.9\textwidth]{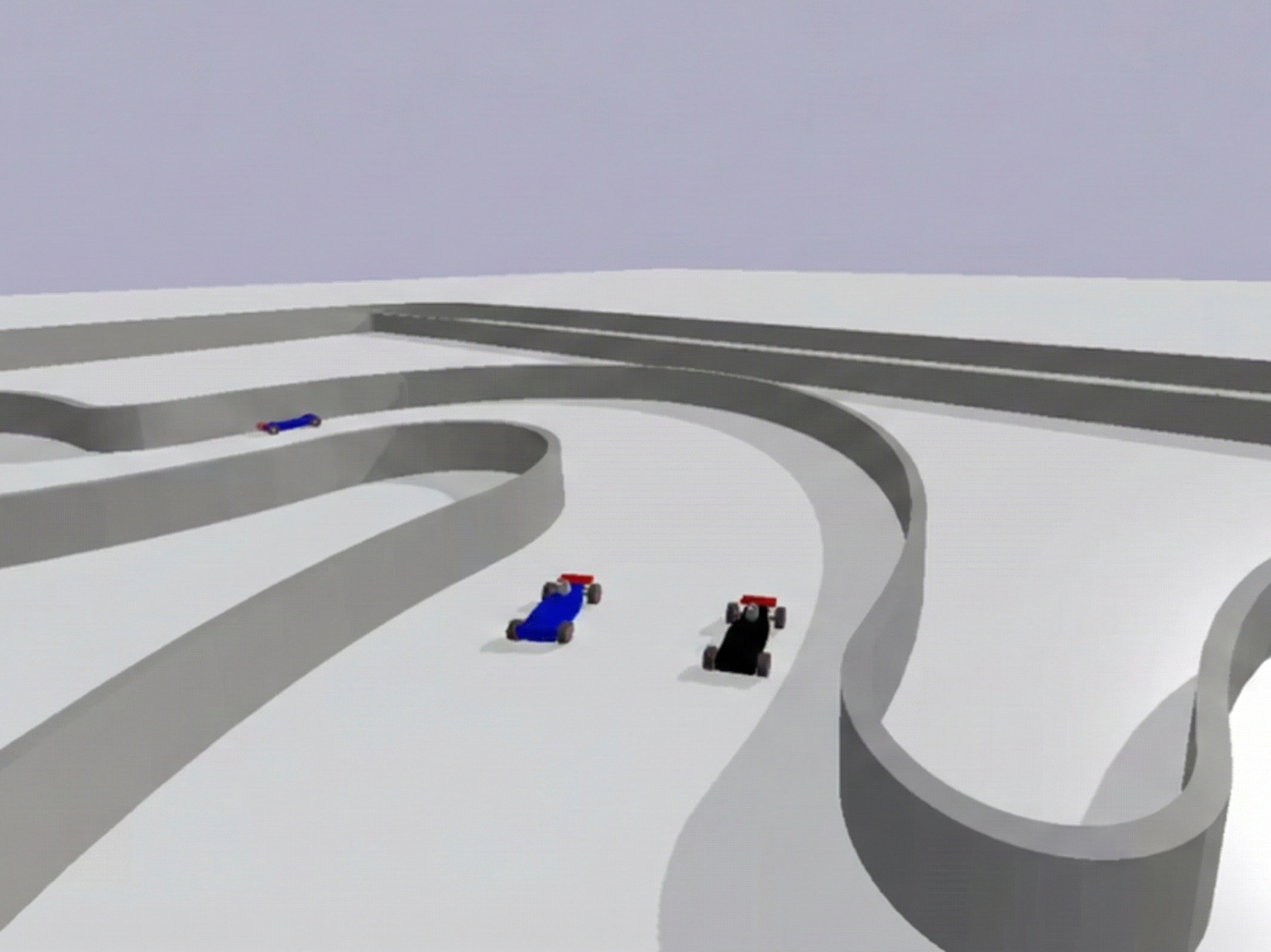}
\subcaption{} \label{racetrack_ot_traj_1}
\end{minipage}%
\begin{minipage}{0.252\textwidth}
\includegraphics[width=0.9\textwidth]{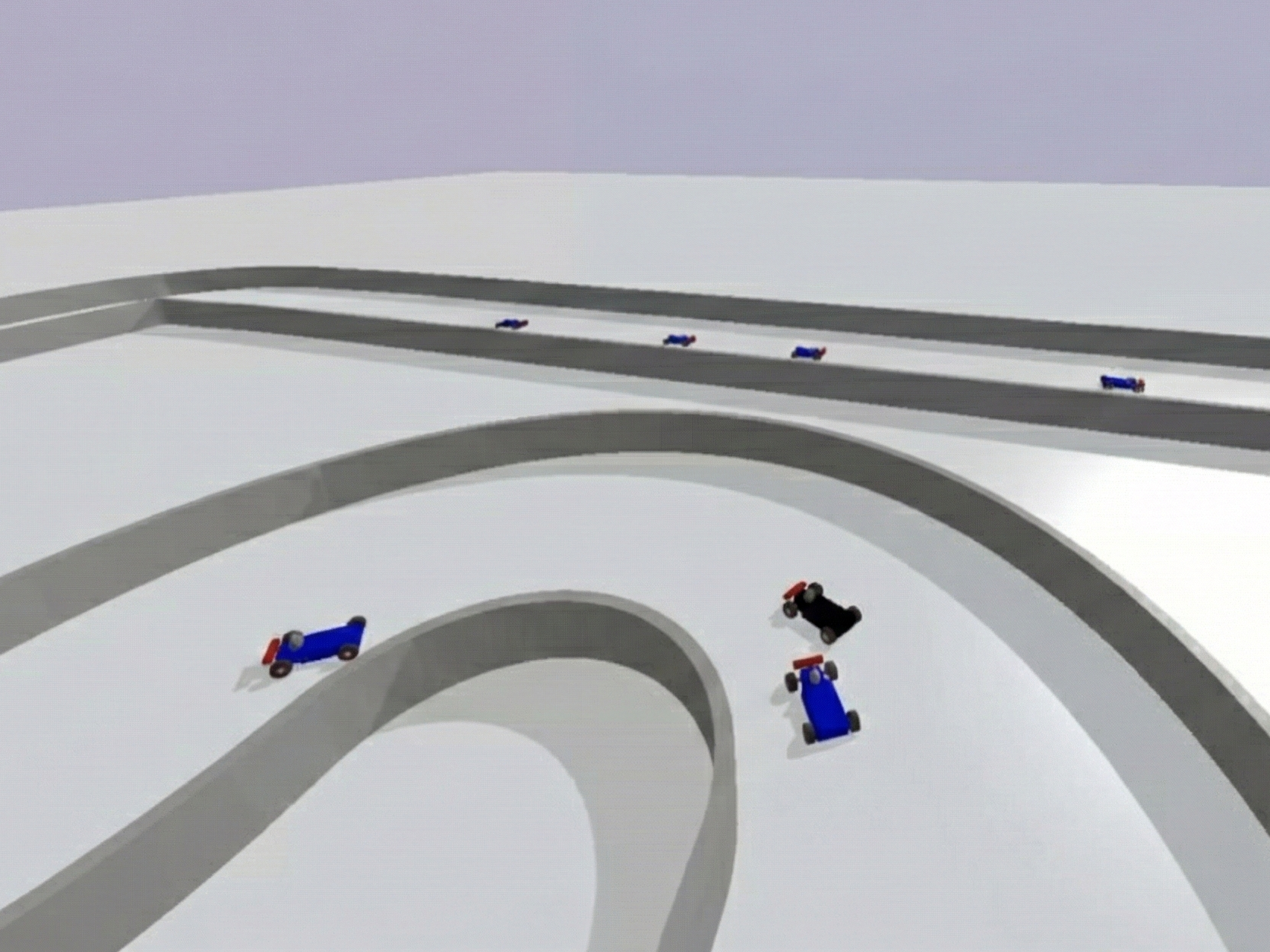}
\subcaption{} \label{racetrack_ot_traj_2}
\end{minipage}
\\
\begin{minipage}{0.252\textwidth}
\includegraphics[width=0.9\textwidth]{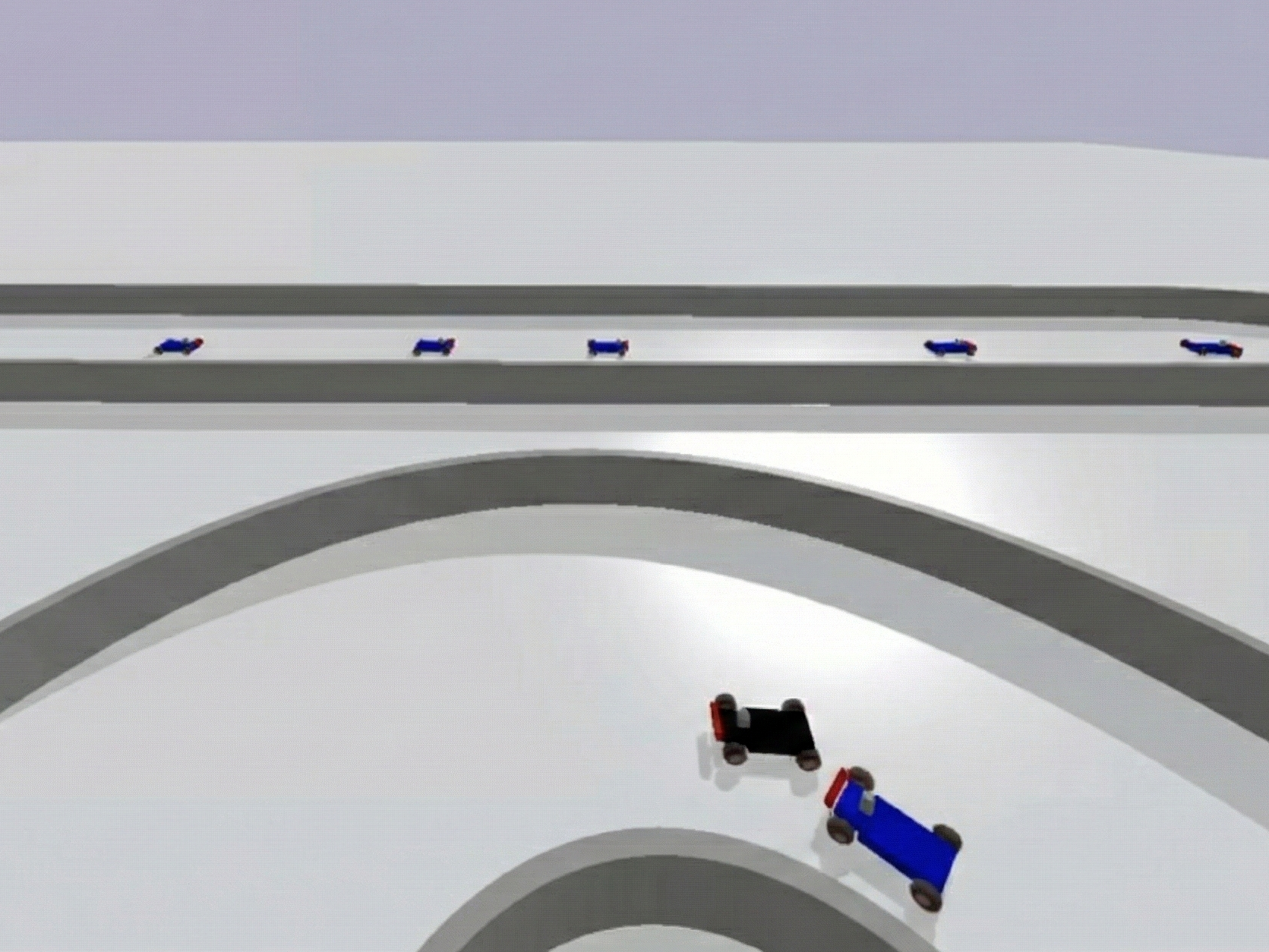}
\subcaption{} \label{racetrack_ot_traj_3}
\end{minipage}%
\begin{minipage}{0.252\textwidth}
\includegraphics[width=0.9\textwidth]{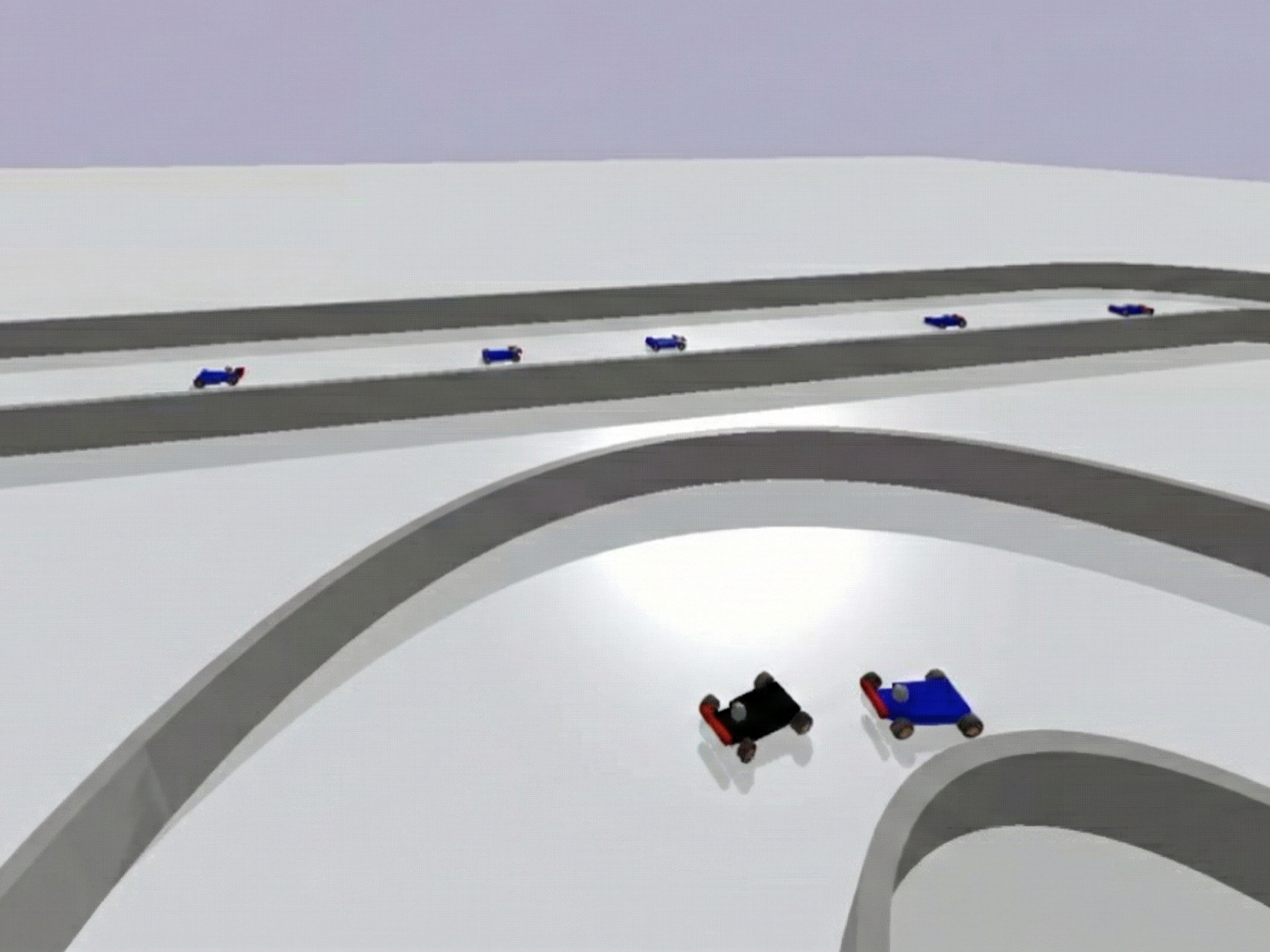}
\subcaption{} \label{racetrack_ot_traj_4}
\end{minipage}
\caption{Example of an overtake motion sequence. (a) The overtaking car switches lane to traverse side-by-side. (b) Ahead before the corner apex on the outside. (c) The obstacle car is cleared and the overtaking car navigates back to the racing line. (d) The overtake is complete.} \label{racetrack_ot_traj}
\end{figure}

The ego agent inferred optimized overtaking trajectories against the dynamic obstacle vehicles, illustrated by the trajectory plots of the overtaking and obstacle cars in Figure \ref{plot_ot_traj} and a motion sequence in Figure \ref{racetrack_ot_traj}, with the same core DRL formulation, with 40,000,000 samples of post-training in the multi-agent environment. 

\subsection{Neural Mechanisms and System Identification}

To elucidate the underlying mechanisms of the end-to-end framework, a comprehensive dissection of the ANN was conducted to map the continuous vehicle controls and evaluate dynamics at the physical handling limits. The hidden layers delineate spatial feature extraction, which compresses the observations to an information-dense vector, and nonlinear kinodynamic execution, with efficient functional bifurcation. Furthermore, system identification validates that the policy maximizes the friction circle by implicitly parameterizing a nonlinear quasi Pacejka tire model.

\begin{figure*}
    \centering
    \begin{minipage}{\textwidth}
    \includegraphics[width=\textwidth]{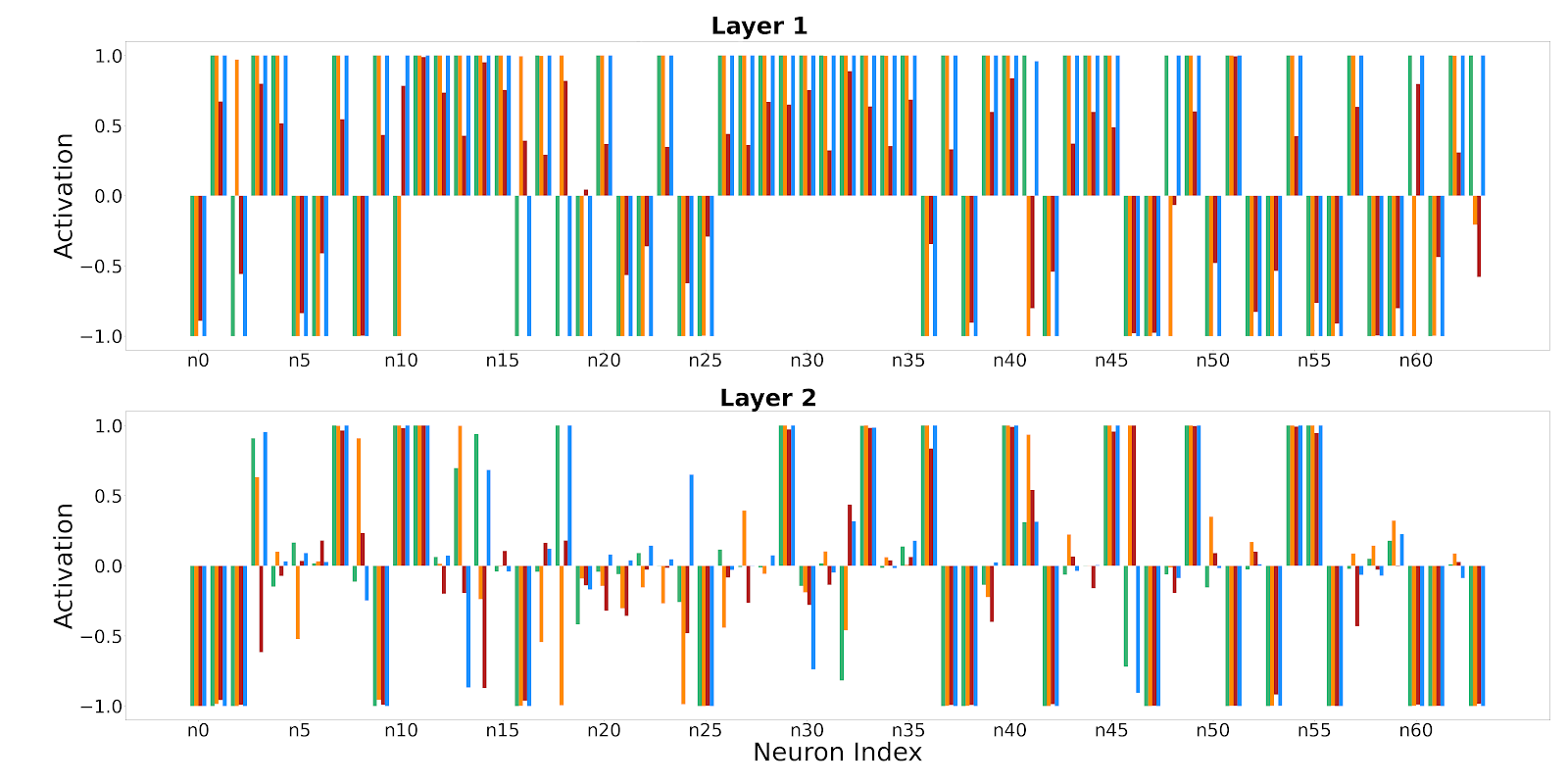}
    \end{minipage}%
    \\
    \begin{minipage}{\textwidth}
    \centering
    \includegraphics[width=0.6\textwidth]{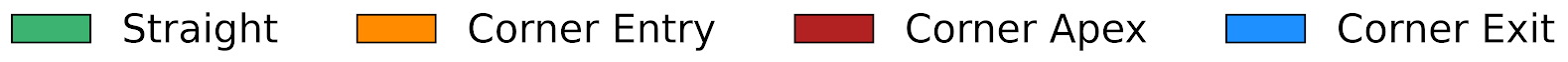}
    \end{minipage}%
    \caption{ANN hidden layer activations.} \label{layer_activations}
\end{figure*}

\textbf{Internal Neural Mechanisms and Control Mapping: }The first hidden layer, $Layer 1$ transforms the high-dimensional observation manifold into a discretized state space. Across the trajectory, this layer operates primarily in a saturated regime, driving $\tanh$ activations to the asymptotes to quantize continuous range data and identify track segments such as straights and corner entries, apexes and exits. Conversely, the second hidden layer, $Layer 2$ maintains an unsaturated balanced distribution across all driving stages, translating $Layer 1$'s discrete spatial states into nonlinear vehicle dynamics considerate continuous controls. We inferred this by analyzing the layer activations segmented into saturation percentage brackets, summarized in Table \ref{table_layer_saturations}, and correlating track features, layer outputs and computed controls. The hidden layer activations at different track segments are illustrated in Figure \ref{layer_activations}.

\begin{table*}[h]
\centering
\caption{Neural Activation Saturation Rates by Driving Stage}
\label{table_layer_saturations}
\resizebox{0.6\textwidth}{!}{
\begin{tabular}[t]{|c|c|c|c|c|c|}
\hline
\textbf{Layer} & \textbf{Stage} & \textbf{Low} & \textbf{Med-Low} & \textbf{Med-High} & \textbf{Saturated} \\
 & & $0 - 25 \%$ & $25 - 50 \%$ & $50 - 75 \%$ & $75 - 100 \%$ \\ \hline
\multirow{4}{*}{1} & Straight & 0.0\% & 0.0\% & 0.0\% & \textbf{100.0\%} \\
& Corner Entry & 1.6\% & 0.0\% & 0.0\% & 98.4\% \\
& Corner Apex & \textbf{3.1\%} & \textbf{32.8\%} & \textbf{28.1\%} & 35.9\% \\
& Corner Exit & 0.0\% & 0.0\% & 0.0\% & \textbf{100.0\%} \\
\hline
\multirow{4}{*}{2} & Straight & \textbf{43.8\%} & 4.7\% & 3.1\% & 48.4\% \\ 
& Corner Entry & 32.8\% & 10.9\% & \textbf{4.7\%} & \textbf{51.6\%} \\
& Corner Apex & 39.1\% & \textbf{12.5\%} & 3.1\% & 45.3\% \\
& Corner Exit & \textbf{43.8\%} & 3.1\% & \textbf{4.7\%} & 48.4\% \\
\hline
\end{tabular}
}
\end{table*}

In the straight and corner exit, $Layer 1$ is polarized, with $100.0\%$ of neurons operating in the highest $75\%$ to $100\%$ saturation bracket, serving as a robust reflex mode. As the vehicle approaches the inner wall during corner entry, preliminary desaturation begins, to prepare the network for precise rotational control, evidenced by $1.6\%$ of activations dropping between $0\%$ and $25\%$. At the corner apex, saturation collapses with 64.1\% of activations distributing into the lower tier brackets, transitioning from a binary output to a high resolution mode, necessary for high-risk maneuvers close to the track boundary, before resaturating at the exit. As the signal propagates to the subsequent hidden layer, the network abstracts these spatial features into definitive physical commands representing steering direction, magnitude, and throttle modulation. Unlike the initial extraction layer, this second layer maintains a balanced, unsaturated distribution across all driving stages.

Specific nodes correspond to distinct physical phenomenon, such as coupled throttle and steering control. Cross-correlation analysis between the hidden layers reveals a dedicated connectivity matrix that implements specific control laws with distinct neural mechanisms.

Neurons 19 and 43 in $Layer 1$, L1-n19 and L1-n43, correlate +0.942 and -0.774  to steering, with the former specializing in left turns, and the latter in right. Each correlates to the specific neurons 36 and 13 in $Layer 2$, L2-n36 and L2-n13. L1-n19 and L2-n36 have an inverse correlation of -0.996, and L1-n43 and L2-n13 correlate by +0.847. L2-n36 correlates -0.945 to steering magnitude, functioning as a steering lock in straights for stable, momentum conserving trajectories. Its activation governs both steering direction and magnitude, where a large positive value induces a hard right turn and vice versa. L2-n13 correlates 0.993 to steering direction, with similar activations, passing a discrete directional boundary to the output layer. Each modulates throttle for stable application of steering with the exception of L2-n13, with L1-n19 correlating by -0.907, L1-n43 by +0.902 and L2-n36 by +0.909. When L2-n36 drops to a negative state to execute a turn for instance, the throttle inherently decreases to trail-brake. Moreover, L2-n14 exhibits a correlation of +0.825 to the throttle, stiffening the steering response as the vehicle accelerates with learned gain scheduling to ensure asymptotic stability at high speeds while retaining agility at low velocities. 

Neurons in $Layer2$ L2-n6, L2-n33 and L2-n57 manage precise rotation at the apex. As the car enters a corner and $Layer 1$ desaturates, the steering lock disengages as its absolute activation drops to $0.84$. Simultaneously, L2-n6, L2-n33 and L2-n57, which remain dormant on straights, exhibit over a $1000\%$ surge in activity, taking over primary control.

\begin{figure}
    \centering
    \begin{minipage}{0.45\textwidth}
    \includegraphics[width=\textwidth]{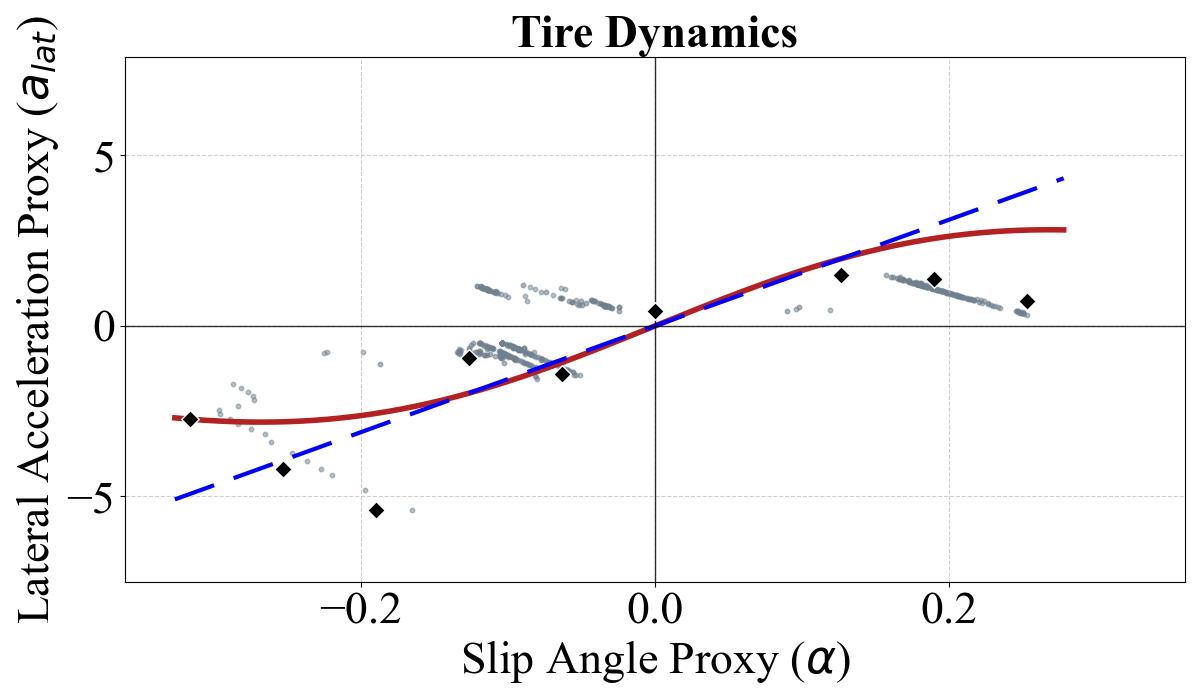}
    \end{minipage}%
    \\
    
    \begin{minipage}{0.5\textwidth}
    \centering
    \includegraphics[width=0.6\textwidth]{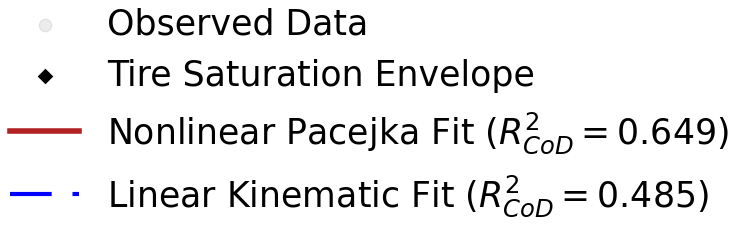}
    \end{minipage}%
    \caption{Tire dynamics: Lateral acceleration across the slip angle range.} \label{plot_tire_model}
\end{figure}

\textbf{Nonlinear Tire Dynamics. } By extracting the 95th percentile limit of adhesion envelope, we evaluated the tire dynamics by fitting standard linear kinematic and empirical Pacejka tire models. The plot of the lateral acceleration proxy against the slip angle proxy is shown in Figure \ref{plot_tire_model}. 

The linear kinematic model fit the adhesion envelope with a coefficient of determination, $R_{CoD}^2 \approx 0.485$. In contrast, the nonlinear Pacejka model yields a greater $R_{CoD}^2 \approx 0.648$. The former does not capture the policy's behavior at the friction limit, whereas the latter better models the tire dynamics with a steep shape factor $C=3.00$ and a flattened saturation plateau corresponding to a peak adhesion limit of $D \approx 2.79g$. Rather than adhering to kinematic predictions, the policy applies over-drive steering commands by holding the slip angle at the boundary of the tire friction circle, to handle the vehicle in the nonlinear saturation region and continuously maximize the grip capacity.

\section{Conclusions}

This paper presented a DRL method for map-free autonomous racing, trained with a non-geometric mimicry, physics-informed reward in simulation over 48 wall clock hours, with zero-shot transfer to hardware, outperforming human demonstrations by 12\% in OOD tracks. The independence of trajectory aided learning enables parameterizing dynamics-optimized overtaking with the same RL formulation, in a multi-agent environment. The policy maximizes the velocity potential of the track geometry from spectral spatial densities, executing nonlinear dynamics considerate continuous controls to maximize the friction circle and carry high momentum. The replacement of an explicit collision penalty with an implicit truncation of the value horizon and the pruning of a non-physical simulator exploit ensured greater OOD generalization and stable transfer to hardware. Nonlinear dynamics were encoded at a low curriculum speed, for policy convergence, and transferred to high-speed domains without intermediate speed scheduling over 15,747 collisions to condense the relations between spatial observations and rigid body dynamics into an efficient MLP with less than 1\% of the computational footprint of SOTA BC and model-based DRL. Through ANN inter-layer correlation analysis, a distinct functional bifurcation was identified, where the first layer compresses observations to extract digitized track features with higher resolution in corner apexes, and the second computes continuous controls from these discrete states with inhibitory and excitatory pathways. Furthermore, system identification of tire dynamics corroborates that the policy implicitly encodes a quasi nonlinear Pacejka tire model to handle the vehicle at the boundary of the tire friction circle to maximize grip and acceleration.

\section*{Acknowledgments}

This work was funded in part by The Commonwealth Cyber Initiative (CCI).

\bibliography{root}
\bibliographystyle{IEEEtran}

\end{document}